%% file: egpaper_arxiv.tex
\documentclass[10pt,twocolumn]{article}

\usepackage{iccv}
\usepackage{times}
\usepackage{graphicx}
\usepackage{amsmath}
\usepackage{amssymb}
\usepackage{booktabs}

\usepackage{comment}
\usepackage{cite}
\usepackage{amsmath,amssymb} 
\usepackage{color}
\usepackage{varwidth}
\usepackage{xcolor}

\usepackage{verbatim}
\usepackage{adjustbox}
\usepackage{array}
\usepackage{bbm}
\usepackage{mathtools}
\usepackage{grffile}
\usepackage{enumitem}
\usepackage{scalerel}
\usepackage[english]{babel}

\usepackage[accsupp]{axessibility}  

\input{math_commands}

\usepackage{algorithm}
\usepackage{algpseudocode}
\usepackage{bbm}
\usepackage{nameref}

\usepackage[pagebackref=true,breaklinks=true,letterpaper=true,colorlinks,bookmarks=false,hypertexnames=false]{hyperref}

\usepackage[capitalize]{cleveref}
\crefname{section}{Sec.}{Secs.}
\Crefname{section}{Section}{Sections}
\Crefname{table}{Table}{Tables}
\crefname{table}{Tab.}{Tabs.}

\iccvfinalcopy 

\def\iccvPaperID{4462} 

\ificcvfinal\fi

\begin{document}

\def\gNsmall{{\scaleto{\gN}{3.5pt}}}
\def\gCsmall{{\scaleto{\gC}{3.5pt}}}
\newcolumntype{x}[1]{>{\centering\let\newline\\\arraybackslash\hspace{0pt}}p{#1}}

\DeclarePairedDelimiterX{\infdivx}[2]{(}{)}{%
  #1\;\delimsize\|\;#2%
}

\title{Learning to Generate Semantic Layouts\\ for Higher Text-Image Correspondence in Text-to-Image Synthesis}

\author{
Minho Park$^*$, \quad Jooyeol Yun$^*$, \quad Seunghwan Choi, \quad Jaegul Choo\\
Korea Advanced Institute of Science and Technology (KAIST)\\
Daejeon, Korea\\
{\tt\small  \{m.park, blizzard072, shadow2496, jchoo\}@kaist.ac.kr}
}

\maketitle
\ificcvfinal\thispagestyle{empty}\fi

\begin{abstract}
    Existing text-to-image generation approaches have set high standards for photorealism and text-image correspondence, largely benefiting from web-scale text-image datasets, which can include up to 5~billion pairs. 
    However, text-to-image generation models trained on domain-specific datasets, such as urban scenes, medical images, and faces, still suffer from low text-image correspondence due to the lack of text-image pairs. 
    Additionally, collecting billions of text-image pairs for a specific domain can be time-consuming and costly.
    Thus, ensuring high text-image correspondence without relying on web-scale text-image datasets remains a challenging task. 
    In this paper, we present a novel approach for enhancing text-image correspondence by leveraging available semantic layouts. 
    Specifically, we propose a Gaussian-categorical diffusion process that simultaneously generates both images and corresponding layout pairs.
    Our experiments reveal that we can guide text-to-image generation models to be aware of the semantics of different image regions, by training the model to generate semantic labels for each pixel. 
    We demonstrate that our approach achieves higher text-image correspondence compared to existing text-to-image generation approaches in the Multi-Modal CelebA-HQ and the Cityscapes dataset, where text-image pairs are scarce. 
    Codes are available in this \href{https://pmh9960.github.io/research/GCDP/}{link}.

\end{abstract}
\footnotetext[1]{\vspace{-0.7cm} * indicates equal contribution.}

\vspace{-0.6cm}
\section{Introduction}
\label{sec:intro}
\vspace{-0.1cm}
Text-to-image generation aims to materialize text descriptions into images, where the main challenge comes from ensuring high image quality and correspondence between input text and output images. 
While texts convey intuitive semantic depictions of images, they often lack detailed spatial descriptions. 
For example, text descriptions such as \emph{``A woman is wearing earrings.''} do not describe where the earrings are located within the image. 
Thus, when a small number of text-image pairs are given, it is challenging for a generative model to learn what part of the image corresponds to which words in the text. 

Overcome this hurdle, recent text-to-image generation approaches~\cite{dalle, dalle2, imagen, ldm} leverage web-scale text-image datasets~\cite{dalle, laion} containing up to 5~billion text-image pairs. 
With access to such data, generative models can fully learn the correspondence between input texts and output images and synthesize photorealistic images while properly reflecting text descriptions.

\begin{figure}[t]
    \centering
    \includegraphics[width=\columnwidth]{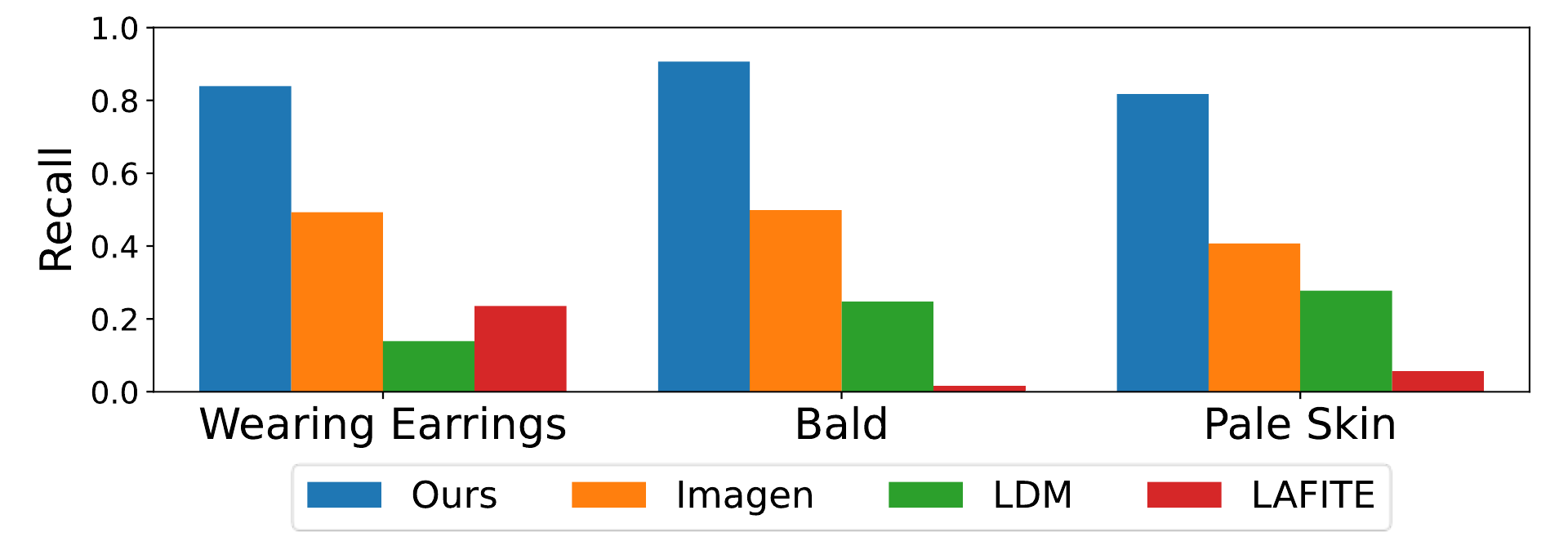}
    \vspace{-0.5cm}
    \caption{Recall of facial attributes specified in the text descriptions. Text-to-image generation approaches trained on a subset of the Multi-Modal CelebA-HQ~\cite{celeba, pggan} often fail to reflect text conditions. Facial attributes are classified with a pretrained attribute classifier~\cite{mldecoder}. 
    }
    \label{fig:intro_semantic_recall}
    \vspace{-0.6cm}
\end{figure}

However, the cost of such large-scale training remains a major obstacle, often requiring weeks of training even with hundreds of GPUs, which limits participation in the subject to only a few researchers.
Moreover, when generating images in a specific domain, such as faces or urban scenes, collecting billions of text-image pairs can be challenging due to the difficulties in collecting images. 
Even with a general-purpose pretrained model, finetuning on datasets with large domain gaps (\eg, urban scenes or medical images) leads to poor image quality and low text-image correspondence.
Recent text-to-image models trained on specific domains often fail to reflect text conditions in the absence of web-scale text-image pairs. 
To examine this issue in data-scarce scenarios, we evaluate text-to-image generation models trained on a subset of the Multi-Modal CelebA-HQ~\cite{pggan, celeba} dataset. 
As shown in \Cref{fig:intro_semantic_recall}, existing models struggle to generate certain attributes specified in the given text conditions. 
Thus, ensuring high text-image correspondence remains a challenge for domain-specific generation.

In this paper, we present a novel approach to achieve high text-image correspondence for domain-specific text-to-image generation by leveraging semantic layouts. 
Rather than solely generating images based on text descriptions, we propose to concurrently generate both images and their corresponding semantic layouts. 
To this end, we design a Gaussian-categorical diffusion process that models the joint distribution of image-layout pairs. 
To the best of our knowledge, this is the first approach to combine Gaussian and categorical diffusion processes into a unified diffusion process. 
By generating semantic labels for each pixel in the image, our generative model can learn the semantics of different parts of the image, allowing it to effectively learn which text descriptions correspond to which locations in the image, even with limited text-image pairs. 

We experiment our approach on subsets of the Multi-Modal CelebA-HQ~\cite{celebamask, celeba} to simulate cases where text-image pairs are limited and semantic layouts are available. 
We also add text descriptions to the Cityscapes dataset~\cite{cityscapes} to evaluate text-to-image generation in complex scenes with multiple objects, where learning text-image correspondence can be challenging. 
Our experiments and analyses reveal that modeling the joint image-layout distribution can effectively facilitate text-to-image generation models to achieve high text-image correspondence when web-scale text-image pairs are unavailable. 
We also demonstrate potential applications of the Gaussian-categorical diffusion models in semantic image synthesis and semantic segmentation, through cross-modal outpainting. 

Our contributions are threefold:
\vspace{-0.15cm}
\begin{itemize}
    \item We define a Gaussian-categorical diffusion process for modeling joint image-layout distributions, which is the first approach to unify two diffusion processes for image-layout generation. 
    \vspace{-0.15cm}
    \item Our experiments reveal that generating image-layout pairs can be a practical alternative to increase text-image correspondence in circumstances where collecting web-scale text-image pairs is infeasible. 
    \vspace{-0.15cm}
    \item We present cross-modal outpainting, which demonstrates that Gaussian-categorical diffusion models are also capable of modeling conditional distributions for semantic image synthesis and semantic segmentation. 
    \vspace{-0.15cm}
\end{itemize}
\vspace{-0.2cm}

\section{Related work}
\vspace{-0.2cm}

\noindent\textbf{Text-to-image generation.}
Text-to-image generation~\cite{attngan, dmgan, xmcgan, lafite} have consistently advanced over the years benefiting from large pretrained text encoders~\cite{t5, dalle} and generative models~\cite{gan, ddpm, dalle}.
Recent approaches~\cite{imagen, glide, dalle2, ldm} tackle zero-shot text-to-image generation by training diffusion-based generative models on web-scale text-image datasets, such as the LAION-5B~\cite{laion} or the DALL-E dataset~\cite{dalle}, which scale from 250M to 5B text-image pairs. 
While zero-shot text-to-image generation can synthesize realistic images given general text descriptions, these approaches heavily rely on the large number of text-image pairs used for training to achieve high text-image correspondence. 
Thus, when these models are trained on specific datasets (\eg, MM CelebA-HQ~\cite{celeba, pggan, celebamask, tedigan}) to generate images within a certain domain, they often fail to satisfy the given text conditions as seen in \Cref{fig:intro_semantic_recall}. 
Collecting enough text-image pairs for a specific domain to ensure high text-image correspondence may be overly expensive since obtaining text descriptions often require human captioning. 
In this paper, we present an alternative approach for enhancing text-image correspondence without additional text-image pairs by leveraging semantic layouts. 

\begin{figure}[t]
    \centering
    \includegraphics[width=\columnwidth]{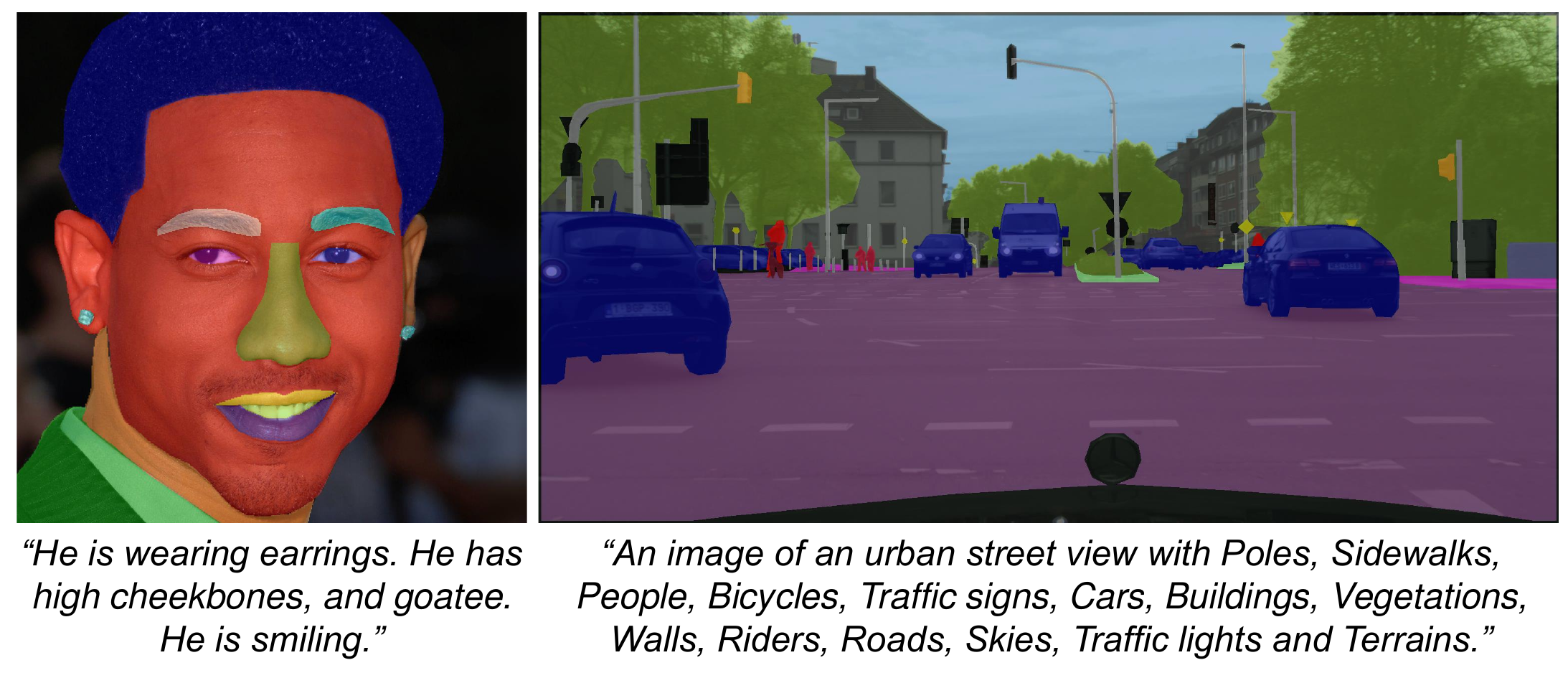}
    \vspace{-0.6cm}
    \caption{Samples of image, text, and layout triplets from the MM CelebA-HQ~\cite{celeba, pggan, celebamask} and the Cityscapes dataset~\cite{cityscapes}.}
    \label{fig:exp_dataset_example}
    \vspace{-0.6cm}
\end{figure}

\noindent\textbf{Generating image-layout pairs.}
Modeling the joint image-layout distribution $p(x,y)$ is an emerging field in image synthesis, where the goal is to generate both the image $x$ and the corresponding semantic layout $y$. 
For the purpose of training semantic segmentation models with strong data augmentation, DatasetGAN~\cite{datasetgan} and DatasetDDPM~\cite{datasetddpm} represent the joint image-layout distribution as a composition of two models: an image generation model $p(x)$ and a classifier $p\cond{y}{x}$. 
During inference, the internal representations of $p(x)$ (\ie, feature maps) are used as inputs of $p\cond{y}{x}$, which then classifies each pixel to obtain an image-layout pair.

On the other hand, SB-GAN~\cite{sbgan} and Semantic Palette~\cite{semanticpalette} discover that joint modeling of the image-layout distribution can be advantageous for generating complex scenes. 
Specifically, they decompose the generation process into two steps, a layout generation step $p(y)$ followed by a conditional image generation step $p\cond{x}{y}$ given the generated layout.
The authors argue that generating layouts with appropriate class proportions can effectively facilitate the scene generation process. 

SemanticGAN~\cite{semanticgan} models $p(x,y)$ with a single GAN~\cite{gan} in the pursuit of semantic segmentation with out-of-domain generalization.
The results demonstrate that images and layouts can exhibit high alignment when generated through a single model. 

In this work, we propose a Gaussian-categorical diffusion process to model $p(x,y)$ with a single diffusion process. 
Our joint image-layout generation model is extended to the text-to-image generation task, where we achieve high text-image correspondence without requiring web-scale text-image datasets.
Specifically, we provide analyses demonstrating that our model is aware of the semantics of the generated image and properly reflects the text conditions.

\begin{figure}[t]
    \centering
    \includegraphics[width=\columnwidth]{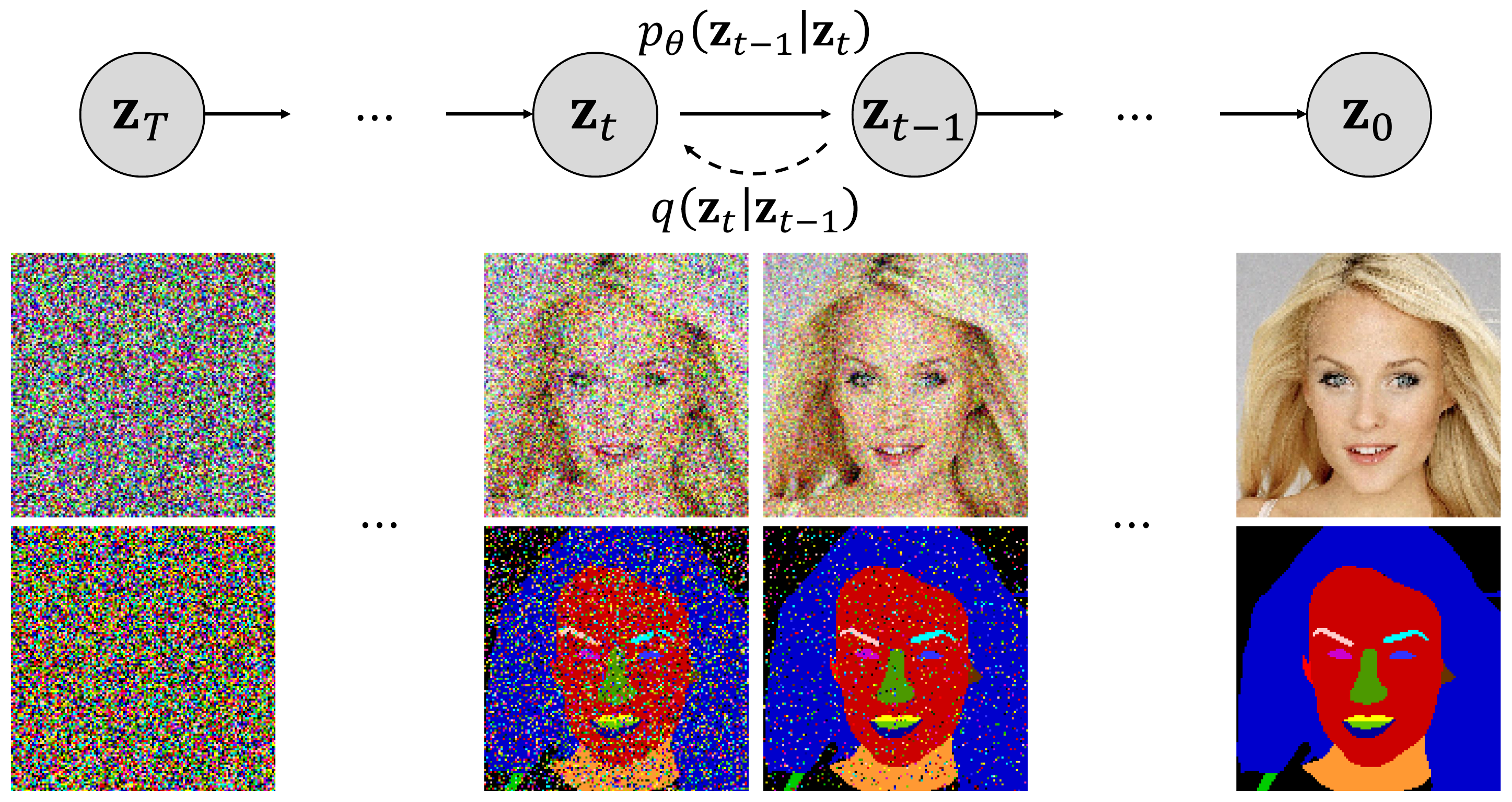}
    \vspace{-0.4cm}
    \caption{Illustration of the Gaussian-categorical diffusion process on the image-layout distribution of MM CelebA-HQ~\cite{celeba, celebamask}. }
    \label{fig:method_joint_diffusion_process}
    \vspace{-0.6cm}
\end{figure}

\noindent\textbf{Diffusion process in continuous and discrete domains.}
Diffusion models~\cite{ddpm, iddpm, ddim, adm} synthesize data $\rvx_0$ in an iterative manner by repeatedly denoising pure noise $\rvx_T$. 
In image generation, the forward noising process $q\cond{\rvx_t}{\rvx_{t-1}}$ and the reverse denoising process $p_\theta\cond{\rvx_{t-1}}{\rvx_t}$ are defined using a predefined noise schedule $\beta_t$, 
\vspace{-0.1cm}
\begin{gather}
    q\cond{\rvx_t}{\rvx_{t-1}} \coloneqq \gN(\rvx_t;\sqrt{1-\beta_t}\rvx_{t-1}, \beta_t\mI),\\
    p_\theta\cond{\rvx_{t-1}}{\rvx_t} \coloneqq \gN(\rvx_{t-1};\vmu_\theta(\rvx_t), \sigma_t^2\mI),
\end{gather}
where $t\in\brkbig{1, 2, ..., T}$. 

Since the true reverse process $q\cond{\rvx_{t-1}}{\rvx_t}$ is intractable, the reverse process is approximated by minimizing the KL divergence with the posterior $q\cond{\rvx_{t-1}}{\rvx_t, \rvx_0}$ with 
\begin{gather}
    L_t = D_{\text{KL}}(q\cond{\rvx_{t-1}}{\rvx_t, \rvx_0} \parallel p_\theta\cond{\rvx_{t-1}}{\rvx_t}).
\end{gather}

To extend diffusion processes to categorical data~\cite{multinomial, d3pm} such as text or semantic labels, a categorical noise is defined for the forward process, and the denoising diffusion process is constructed in a similar manner. 
For instance, Hoogeboom~\etal~\cite{multinomial} defines a categorical noise as
\begin{gather}
    q\cond{\rvx_t}{\rvx_{t-1}} \coloneqq \gC(\rvx_t;(1-\beta_t)\rvx_{t-1} + \beta_t/K),\\
    p_\theta\cond{\rvx_{t-1}}{\rvx_t} \coloneqq \gC(\rvx_{t-1};\mTheta_\theta(\rvx_t)),
\end{gather}
where $\gC$ denotes a categorical distribution, $K$ is the number of categories, and $\mTheta$ is the probability mass function (PMF) of the categorical distribution.

The key idea for defining a diffusion process in a certain distribution is to define a forward noising process $q\cond{\rvx_t}{\rvx_{t-1}}$ and derive a posterior $q\cond{\rvx_{t-1}}{\rvx_t, \rvx_0}$.
In the following section, we define the forward and reverse processes of the Gaussian-categorical distribution, which can model the joint distribution of image-layout pairs.

\begin{figure}[t]
    \centering
    \includegraphics[width=0.8\columnwidth]{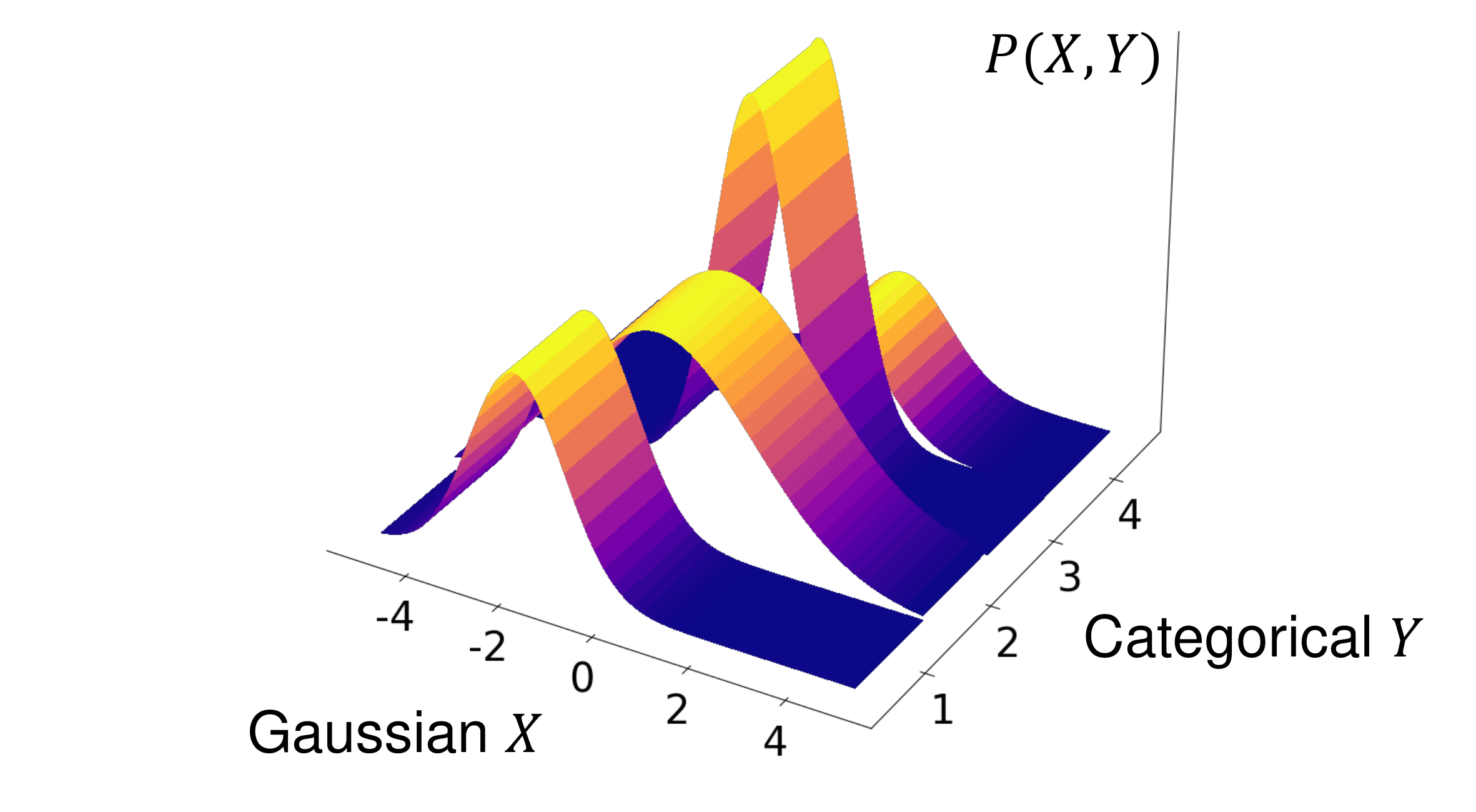}
    \caption{Visualization of a Gaussian-categorical distribution with a single variable ($N=1$, $M=1$, $K=4$, and $S=4$). }
    \label{fig:method_gaussian_categorical}
    \vspace{-0.4cm}
\end{figure}

\section{Method}
\vspace{-0.2cm}

\subsection{Gaussian-categorical distribution}
In this section, we define the joint distribution of the Gaussian variable $X$ and categorical variable $Y$.
We parameterize the Gaussian-categorical distribution as follows,
\vspace{-0.2cm}
\begin{gather}
\label{eq:method_dims}
    (X, Y) \sim \gN\gC \brk{\rvx, \rvy; \vmu, \mSigma, \mTheta},\\
    X=\brkbig{X_1, X_2, ... , X_N}\in\R^N,\nonumber\\
    Y=\brkbig{\emY_1, \emY_2, ... , \emY_M}\in\brkmid{1, 2, ..., K}^M \subset \R^M,\nonumber\\        
    \vmu \in \R^{S \times N}, \mSigma \in \R^{S\times N\times N}, \mTheta \in \R^{M\times K}.\nonumber
\end{gather}
Here, $\vmu, \mSigma$ are the mean and variance of the Gaussian distribution, and $\mTheta$ is the probability mass function (PMF) of the categorical distribution. 
Also, $K$ is the number of possible states for $Y_i$ and $S=K^M$ is the total number of states of $Y$.
It is worth noting that the dimensions of $\vmu$ and $\mSigma$, which indicates that there is a Gaussian mean and variance for all possible categorical states in $Y$.

The joint distribution of two random variables can be written as a product of a conditional and marginal distribution. 
Therefore, we can also express the Gaussian-categorical distribution as
\begin{gather}
    \gN\gC(\rvx, \rvy; \vmu, \mSigma, \mTheta) = \gC(\rvy; \mTheta) \cdot \gN(\rvx; \vmu_\rvy, \mSigma_\rvy) \\
    \vmu_\rvy \in \R^{N}, \mSigma_\rvy \in \R^{N\times N}. \nonumber
\end{gather}
The probability density function (PDF) can be written as a weighted Gaussian distribution for each unique $\rvy \in \brkmid{1, 2, ..., K}^M$ as
\begin{align}
    \gN\gC(\rvx, \rvy; \vmu, \mSigma, \mTheta) = &\brk{ \prod_{i=1}^M \mTheta_{\scaleto{i}{6pt}\,,\, \scaleto{\rvy}{6pt}_{\scaleto{i}{4.5pt}}} } \brk{2\pi}^{-\frac{N}{2}} \abs{\mSigma_\rvy}^{-\frac{1}{2}} \nonumber \\
    &\exp{ \Big( -\frac{1}{2} (\rvx - \vmu_\rvy)^\top \mSigma^{-1}_\rvy \brk{\rvx - \vmu_\rvy} \Big) },
\end{align}
where $\mTheta_{\scaleto{i}{6pt}\,,\, \scaleto{\rvy}{6pt}_{\scaleto{i}{4.5pt}}}$ denotes the probability of $Y_i = \rvy_i$, and $\vmu_\rvy$, $\mSigma_\rvy$ indicates the mean and variance corresponding to state $\rvy$, respectively.

\subsection{Gaussian-categorical diffusion process}
Similar to the diffusion process for images, we define our reverse process of image-layout distributions as a Gaussian-categorical transition with a Markov property.
Specifically, we define the transition probability $p_\theta\cond{\rvz_{t-1}}{\rvz_t}$ as
\begin{equation}
    p_\theta \cond{\rvz_{t-1}}{\rvz_t} \coloneqq \gN\gC \brk{\rvz_{t-1} ; \vmu_\theta (\rvz_t), \mSigma_\theta (\rvz_t), \mTheta_\theta (\rvz_t)},
\end{equation}
where $\rvz$ represents the tuple $(\rvx,\rvy)$ for simplicity.

We define the forward process of image-layout pairs $\rvz_0$ under the Markov assumption as
\begin{equation}
\label{eq:method_forward}
    q \cond{\rvz_{t}}{\rvz_{t-1}} \coloneqq \gN\gC \Bigl( \rvz_{t} ; \bigl[ \vmu_{t|t-1} \bigr]_{\times S}, \bigl[ \mSigma_{t|t-1} \bigr]_{\times S}, \mTheta_{t|t-1} \Bigl),
\end{equation}
\vspace{-0.5cm}
\begin{align*}
    \vmu_{t|t-1} &\coloneqq \sqrt{1 - \beta^\gNsmall_t} \rvx_{t-1},\\
    \mSigma_{t|t-1} &\coloneqq \beta^\gNsmall_t \mI, \\
    \mTheta_{t|t-1} &\coloneqq (1 - \beta^\gCsmall_t) \rvy_{t-1} + \beta^\gCsmall_t / K,
\end{align*}
where $\beta^\gCsmall$ and $\beta^\gNsmall$ are predefined noise schedules. 
We use the notation $[ \rvv ]_{\times S}$ to indicate row-wise duplication of a vector $\rvv$ (\ie, $[ \rvv, \rvv, ..., \rvv]^T$).

Intuitively, the forward process is defined as independently applying the Gaussian and categorical noises following a normal distribution $\gN(\mathbf{0}, \mI)$ and a categorical distribution with uniform probability $\gC(1/K)$, according to predefined noise schedules $\beta^\gNsmall, \beta^\gCsmall$. 
Given a large $T$ and appropriate noise schedules, the forward process leads to an isotropic Gaussian distribution and a uniform categorical distribution at the final state $\rvz_T$.

\begin{figure*}[t]
    \centering
    \includegraphics[width=\textwidth]{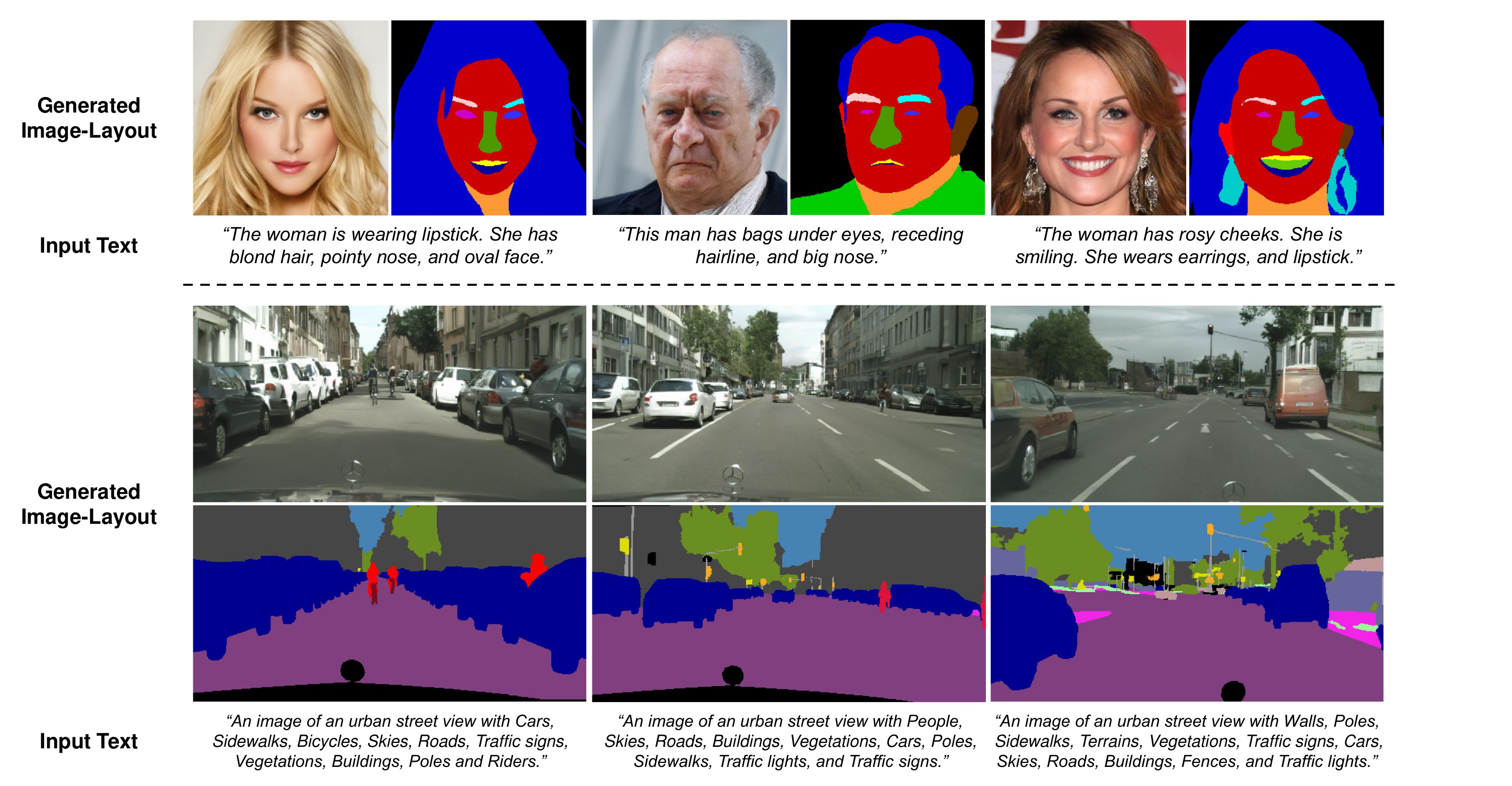}
    \vspace{-0.6cm}
    \caption{Examples of text-guided generation of image-layout pairs from the Gaussian-categorical diffusion trained on MM CelebA-HQ-100~\cite{celeba, pggan} and Cityscapes~\cite{cityscapes}. The text descriptions on the bottom are given as conditions to generate the image-label pairs.}
    \label{fig:exp_quali}
    \vspace{-0.4cm}
\end{figure*}

With $\alpha_t \coloneqq 1 - \beta_t$ and $\bar{\alpha}_t \coloneqq \prod_{s=1}^t \alpha_s$, we can derive a forward process to an arbitrary timestep as
\begin{equation}
    q \cond{\rvz_t}{\rvz_0} = \gN\gC \Bigl( \rvz_{t} ; \bigl[ \vmu_{t|0} \bigr]_{\times S}, \bigl[ \mSigma_{t|0} \bigr]_{\times S}, \mTheta_{t|0} \Bigl),
\end{equation}
\vspace{-0.5cm}
\begin{align*}
    \vmu_{t|0} &\coloneqq \sqrt{ \bar{\alpha}^\gNsmall_t } \rvx_{0},\\
    \mSigma_{t|0} &\coloneqq (1 - \bar{\alpha}^\gNsmall_t) \mI, \\
    \mTheta_{t|0} &\coloneqq (1 - \bar{\alpha}^\gCsmall_t) \rvy_{0} + \bar{\alpha}^\gCsmall_t / K.
\end{align*}

Finally, using Bayes theorem, we can derive the posterior $q\cond{\rvz_{t-1}}{\rvz_t, \rvz_0}$, which is summarized into the following form of a Gaussian-categorical distribution
\vspace{-0.2cm}
\begin{align}
    q \cond{\rvz_{t-1}}{\rvz_t, \rvz_0} = \gN\gC \Bigl( \rvz_{t-1}; \bigl[\widetilde{\vmu}_t \bigr]_{\times S}, \bigl[ \widetilde{\mSigma}_t \bigr]_{\times S}, \widetilde{\mTheta}_t \Bigr), \hspace{-1pt}
\end{align}
\vspace{-0.5cm}
\begin{gather*}
    \widetilde{\vmu}_t \coloneqq \frac{\sqrt{\bar{\alpha}^{\gNsmall}_{t-1}}\beta_t^{\gNsmall}}{1 - \bar{\alpha}^{\gNsmall}_t} \rvx_0 + \frac{\sqrt{\alpha^{\gNsmall}_t}(1-\bar{\alpha}^{\gNsmall}_{t-1})}{1 - \bar{\alpha}^{\gNsmall}_t} \rvx_t,\\
    \widetilde{\mSigma}_t \coloneqq \bigl( (1 - \bar{\alpha}^\gNsmall_{t-1}) \beta^\gNsmall_t / ( 1 - \bar{\alpha}^\gNsmall_{t}) \bigr) \mI,\\
    \widetilde{\mTheta}_t \coloneqq Z [\alpha^\gCsmall_t \rvy_t + (1 - \alpha^\gCsmall_t)/K] \odot [\bar{\alpha}^\gCsmall_t \rvy_0 + (1 - \bar{\alpha}^\gCsmall_{t-1}) / K],
\end{gather*}
\vspace{-0.1cm}
where $Z$ is a normalizing constant and $\odot$ is the element-wise product. 
Detailed proofs for each step are provided in \hyperref[sec:proofs]{A.1}.

Note that parameters $\vmu$ and $\mSigma$ of the posterior are expressed in terms of $\widetilde{\vmu}_t \in \R^{N}$ and $\widetilde{\mSigma}_t \in \R^{N\times N}$, which have a reduced dimensions than the original parameters in \Cref{eq:method_dims}. 
This is due to the definition in \Cref{eq:method_forward}, where the Gaussian noise is applied independently of the categorical variable.

We can write the variational lower bound (VLB) as 
\vspace{-0.1cm}
\begin{gather}
    L_{\text{VLB}} \coloneqq L_0 + L_1 + L_2 + ... + L_T,\\
    L_0 \coloneqq -\log p_\theta\cond{\rvz_0}{\rvz_1},\\
    L_{t-1} \coloneqq D_{KL}(q \cond{\rvz_{t-1}}{\rvz_t, \rvz_0} \parallel p_\theta\cond{\rvz_{t-1}}{\rvz_t}),\label{eq:method_kl}\\
    L_T \coloneqq D_{KL}(q\cond{\rvz_T}{\rvz_0}\parallel p_\theta(\rvz_T)).
\end{gather}
Since the posterior $q \cond{\rvz_{t-1}}{\rvz_t, \rvz_0}$ is parameterized by $\widetilde{\vmu}_t$ and $\widetilde{\mSigma}_t$, we can also re-parameterize $p_\theta$ as
\begin{gather}
    \hspace{-4pt}
    p_\theta\cond{\rvz_{t-1}}{\rvz_t} \coloneqq \gN\gC(\rvz_{t-1}; [\widetilde{\vmu}_\theta(\rvz_t)]_{\times S}, [\widetilde{\mSigma}_\theta(\rvz_t)]_{\times S}, \mTheta_\theta),\hspace{-4pt}\\
    \widetilde{\vmu}_\theta(\rvz_t) \in \R^{N}, \widetilde{\mSigma}_\theta(\rvz_t) \in \R^{N\times N}, \mTheta_\theta \in \R^{M\times K},
\end{gather}
Thus, we can predict a reduced number of parameters to minimize the KL divergence term in \Cref{eq:method_kl},
\begin{align}
    &D_{KL} ( q\cond{\rvz_{t-1}}{\rvz_t, \rvz_0} \parallel p_\theta \cond{\rvz_{t-1}}{\rvz_t} ) \\
    &= \E_q \Bigl[ \frac{1}{2\sigma_t^2} \norm{\widetilde{\vmu}_t - \widetilde{\vmu}_\theta (\rvz_t)}^2 \Bigr] + D_{\text{KL}} ( \widetilde{\mTheta}_t \parallel \mTheta_\theta(\rvz_t) ) + C,
    \nonumber
\end{align}
\vspace{-0.1cm}
where $C$ is a constant irrelevant to learnable parameters $\theta$. 
$L_0$ is directly minimized through a closed-form solution and $L_T$ does not involve any learnable parameters.

\subsection{Architectural design}
In order to treat image-layout pairs as a single data sample, we embed the semantic layouts (\ie, one-hot vectors) into 3-channel vectors via learnable parameters and concatenate them with images along the channel dimension ($\rvz \in \R^{N\times N\times6}$). 
We adopt the U-Net~\cite{iddpm} and the Efficient U-Net~\cite{imagen} following existing diffusion models and modify the input/output channels for image-layout input/outputs. 
For text conditioning, we utilize the T5-L~\cite{t5} text encoder and condition the U-Net model similarly to Imagen~\cite{imagen}. 

We follow the cascaded diffusion~\cite{cdm} framework to generate high-resolution image-layout pairs, which involves a sequence of an image generation model followed by a super-resolution model. 
We find that resizing layouts to a small resolution (\eg, $64\times 64$) often damages the integrity of semantic labels due to nearest-neighbor sampling on extreme scales. 
Thus, we generate $128\times128$ resolution images and then upsample to $256\times256$ resolution with a Gaussian-categorical super-resolution model. 
The super-resolution model upsamples both images and layouts following the Gaussian-categorical diffusion. 
We adopt the classifier-free guidance on both the generation model and the super-resolution model. 

\vspace{-0.4cm}
\section{Experiments}

\subsection{Text-image datasets}
\noindent\textbf{Multi-Modal CelebA-HQ.} MM CelebA-HQ~\cite{celeba, pggan, tedigan} is a collection of different annotations for the 30,000 images in the CelebA-HQ dataset~\cite{pggan, celeba}, including text descriptions, face attribute labels, and part-level segmentation labels. 
Part-level segmentation labels consist of 19 different classes ($K\hspace{-4pt}=\hspace{-4pt}19$) including all facial components and accessories. 
To train the Gaussian-categorical diffusion model, we use both the segmentation labels and the text descriptions provided in the dataset.
We also construct subsets of the data, MM CelebA-HQ-25 and MM CelebA-HQ-50, by randomly selecting $25\%$ and $50\%$ of the images, respectively, to simulate data-scarce scenarios. 
We train and evaluate our models on $256\times256$ resolution images.

\noindent\textbf{Cityscapes.} Cityscapes~\cite{cityscapes} is an urban scene dataset with 3475 image-layout pairs of complex scenes containing multiple objects, including 20 different semantic classes ($K\hspace{-2pt}=\hspace{-2pt}20$). 
To add text descriptions to each image, we list the class names in the following format:
\vspace{-0.2cm}
\begin{equation*}
    \text{``An image of an urban scene with \{\textit{classes}\}.''}
\vspace{-0.2cm}
\end{equation*}
where \emph{classes} are the unique class names in the corresponding semantic layout.
The Cityscapes dataset presents a challenging domain for generating realistic images due to the limited number of available images and the diverse object locations in urban scenes.
Since Cityscapes images have a unique aspect ratio of $2\hspace{-2pt}:\hspace{-2pt}1$, we generate $512\times256$ resolution images.
We include example text-image pairs in \Cref{fig:exp_dataset_example}.

\begin{figure*}[t]
    \centering
    \includegraphics[width=\textwidth]{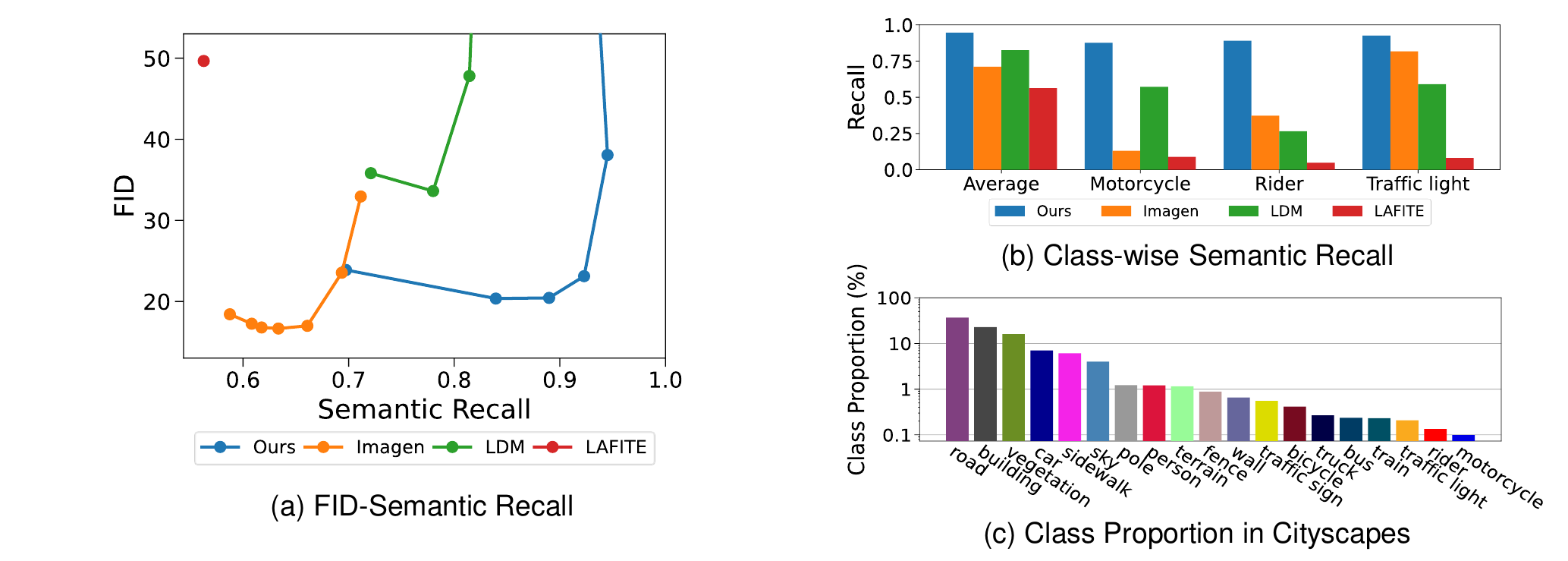}
        \caption{(a) FID-Semantic Recall trade-off in the Cityscapes dataset. (b) Semantic Recall for minor classes. Semantic Recall is measured using the HRNet-w48~\cite{hrnet} model. (c) Proportion of each semantic class in the entire Cityscapes dataset. Class proportion is compared in log-scale for visibility.}
    \label{fig:exp_cityscapes}
    \vspace{-0.4cm}
\end{figure*}

\vspace{-0.2cm}
\subsection{Implementation details}
For synthesizing image-layout pairs, $N$ and $M$ are equally set to the number of pixels in the image. 
Although the Gaussian-categorical diffusion process allows different noise schedules $\beta^\gNsmall$ and $\beta^\gCsmall$ for images and layouts, we set both schedules to the cosine schedule~\cite{iddpm}. 
We provide experiments on the effect of different noise schedules for $\beta^{\gCsmall}$ in the supplementary section. 
We set $T=1000$ and sample with 100 timesteps using the accelerated sampling technique~\cite{ddim}. 

\vspace{-0.3cm}
\subsection{Evaluating text-to-image generation}
\vspace{-0.2cm}
Text-to-image generation models are evaluated from two perspectives, image fidelity and text-image correspondence. 
We use the Fr\'{e}chet Inception Distance (FID)~\cite{fid} to measure the image fidelity. 
After the release of CLIP~\cite{clip}, the CLIP score~\cite{clipscore} is often used to evaluate text-image correspondence for text-to-image generation. 
However, the CLIP score is known to have poor generalization abilities~\cite{clip} when evaluating scenes with large domain gaps (\ie, Cityscapes) and also lacks interpretability in terms of understanding what element in the image causes a low or high CLIP score. 
In order to compensate for this drawback, we propose \emph{Semantic Recall} to precisely measure the text-image correspondence for Cityscapes generation.

\noindent\textbf{Semantic Recall.} The Semantic Recall is analogous to the Semantic Object Accuracy (SOA)~\cite{soa}, which evaluates the generation of specific objects in text-to-image generation by utilizing pretrained object detectors. 
In our work, we use a pretrained semantic segmentation model~\cite{hrnet} to detect the presence of classes described in text conditions. 
We determine that a class is \emph{detected} in a generated image if it appears in the segmentation layout. 
The ground-truth classes for each image are identified by searching for class names in text descriptions. 
For example, an image generated with the text description \emph{``An urban scene with cars, roads, and traffic signs.''}, would be evaluated with the existence of \emph{cars}, \emph{roads}, and \emph{traffic signs}.
Therefore, we compute the Semantic Recall as the average ratio of correctly detected classes in the generated image to the total number of classes in the ground-truth layouts,
\begin{equation*}
    \frac{1}{\mid \mathcal{G}\mid}\sum_{x_i, y_i \in \mathcal{G}}\frac{\mid \text{Classes in }F(x_i) \cap \text{ Classes in } y_i\mid}{\mid \text{Classes in }y_i\mid},
\end{equation*}
where $\mathcal{G}$ is the set of generated image-layout pairs $(x_i, y_i)$ and $\mid\hspace{-2pt}\cdot\hspace{-2pt}\mid$ indicates the cardinality of a given set. 
$F(\cdot)$ is the pretrained semantic segmentation model~\cite{hrnet}.

\noindent\textbf{Baselines.} We compare our approach with state-of-the-art performing diffusion-based models, Imagen~\cite{imagen} and the latent diffusion model (LDM)~\cite{ldm}.
We also train a high-performing GAN-based approach Lafite~\cite{lafite} trained on MM CelebA-HQ and Cityscapes. 
For training LDM, we utilize the pretrained autoencoder from the Stable Diffusion project. 
Diffusion-based approaches utilize the classifier-free guidance~\cite{cfg} to control the performance trade-off between text-image correspondence and image fidelity. 
Thus, for these approaches, we sweep the guidance scale until the text-image correspondence measures saturate and report all FID-Semantic Recall or FID-CLIP score pairs.

\noindent\textbf{Evaluation on Cityscapes.} For the Cityscapes dataset~\cite{cityscapes}, we report the FID and Semantic Recall performance trade-off and also provide detailed recall scores for each class in \Cref{fig:exp_cityscapes}. 
Given the small number of text-image pairs (3475 pairs), existing text-to-image models face challenges in learning the text-image correspondence and achieving high text-image correspondence. 
However, the Gaussian-categorical diffusion effectively generates complex Cityscapes scenes while maintaining high Semantic Recall even with limited data. 
Additionally, the model achieves high recall rates for minor classes, such as the \emph{bicycle} or the \emph{motorcycle} class, which only constitute a small portion of the dataset. 
This indicates that generating semantic labels for each pixel facilitates the model to establish high text-image correspondence, especially for underrepresented classes. 

\begin{figure*}[t]
    \centering
    \includegraphics[width=\textwidth]{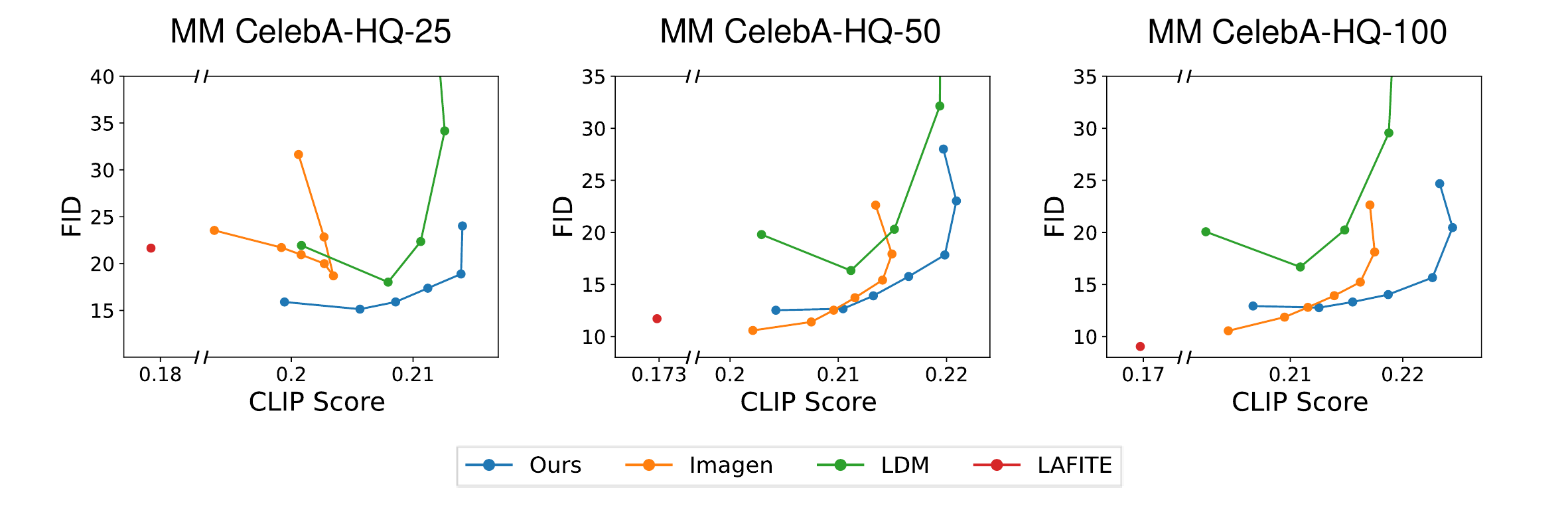}
    \vspace{-0.4cm}
    \caption{FID-CLIP score pairs for text-to-image generation models on different subsets of the MM CelebA-HQ dataset. The CLIP scores are measured with the ViT-L/14-336 model. The guidance scale is swept starting from 1 until saturation.}
    \label{fig:exp_fid_clip_celeba}
    \vspace{-0.4cm}
\end{figure*}

\noindent\textbf{Evaluation on MM CelebA-HQ.} We further evaluate our method on the MM CelebA-HQ-25, 50, and 100, and report the FID-CLIP scores for each dataset. 
As shown in \Cref{fig:exp_fid_clip_celeba}, the Gaussian-categorical diffusion consistently outperforms existing text-to-image approaches at datasets with varying numbers of text-image pairs, exhibiting low FIDs and a high CLIP scores. 
We provide qualitative results of the Gaussian-categorical diffusion in \Cref{fig:exp_quali} and also compare the results with existing approaches in the supplementary material. 

\begin{figure}[t]
    \centering
    \includegraphics[width=\columnwidth]{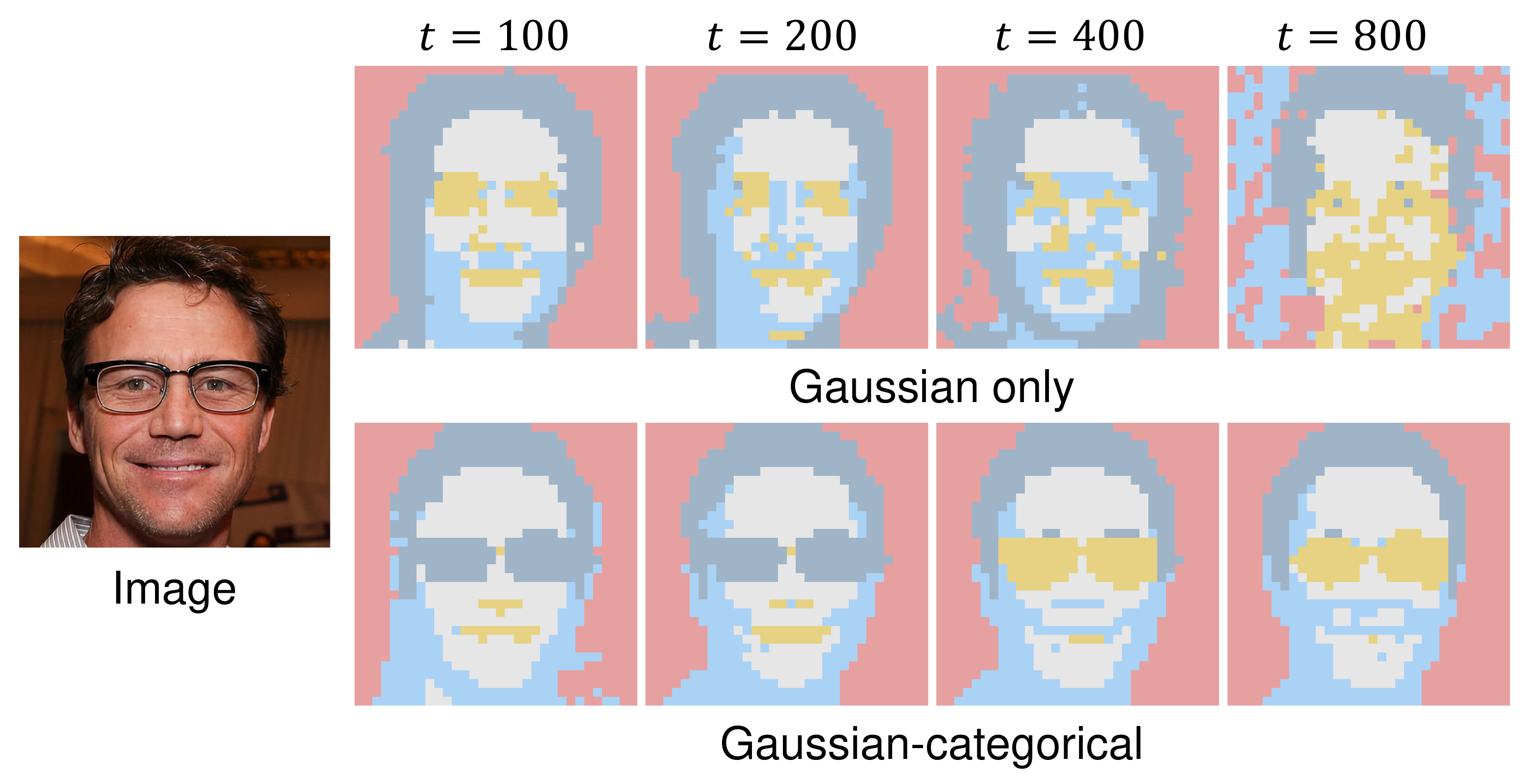}
    \caption{Visualization of clustering results between the internal features of the Gaussian-categorical diffusion and the Gaussian diffusion.}
    \label{fig:exp_clustering}
    \vspace{-0.4cm}
\end{figure}

\subsection{Analyzing the internal representations}
In order to visualize the advantages of jointly generating image-layout pairs, we train a Gaussian diffusion model which generates images without corresponding semantic layouts. 
Then, we collect the internal features from the two models at different timesteps and cluster the features in an unsupervised manner with K-means clustering. 
As shown in \Cref{fig:exp_clustering}, the internal features of the Gaussian-categorical model form distinct clusters that correspond to different facial regions. 
Specifically, the internal features of the Gaussian-categorical diffusion model form clusters even in the early stages of generation ($t\hspace{-0.1cm}=\hspace{-0.1cm}800$), correctly distinguishing hair, glasses and the background region. 

The results reveal that the Gaussian-categorical diffusion model is highly aware of the semantics of the image during the generation process. 
This characteristic is advantageous in scenarios where a generative model needs to learn how to match specific parts of the image with corresponding input text descriptions, as the model is capable of understanding the semantic structure of the image.
As such, training a Gaussian-categorical diffusion is a promising approach for achieving high correspondence between text descriptions and image pixels, particularly when there is a scarcity of text-image pairs available.

\subsection{Image-layout fidelity and alignment}
In this section, we evaluate whether generated images and layouts closely model the real distribution, and whether the generated pairs are semantically aligned. 
Following Semantic Palette~\cite{sbgan, semanticpalette} we evaluate the image-layout alignment using the mean intersection over union (mIoU) between the generated layouts and the segmentation labels predicted by a pretrained HRNet~\cite{hrnet}. 
Additionally, we use the Fr\'{e}chet Segmentation Distance (FSD)~\cite{fsd}, which replaces the Inception-V3~\cite{inception} features in the FID score~\cite{fid} to pixel counts for each class, to evaluate the quality of generated layouts. 
Similar to the FID score, a low FSD indicates that the class distributions are close to the real distribution.

\begin{table}[t]
\centering
\begin{tabular}{@{}x{3.2cm}x{1.2cm}x{1.2cm}x{1.2cm}@{}}
\toprule
Methods                                 & FID $\downarrow$ & mIoU $\uparrow$ & FSD $\downarrow$\\ \midrule
GANformer~\cite{ganformer}              &24.86&   -   & 481.5\\
DatasetDDPM~\cite{datasetddpm}          &55.38& 33.88 & 90.31\\
Semantic Palette~\cite{semanticpalette} &52.13& 53.17 & 48.29\\ \midrule
Ours                                    &\textbf{20.36}& \textbf{65.80} & \textbf{42.22}\\ \bottomrule
\end{tabular}
\vspace{+0.1cm}
\caption{Image-layout alignment and FID of different Image-layout generation approaches for scene generation in the Cityscapes~\cite{cityscapes} dataset.}
\vspace{-0.4cm}
\label{tab:exp_miou}
\end{table}

We compare our results with existing unconditional image-layout generation approaches~\cite{semanticpalette, datasetddpm} on the Cityscapes dataset. 
Additionally, we introduce a simple baseline (\ie, GANformer~\cite{ganformer}) for image-layout generation, in which we generate images using a well-trained unconditional image generation model~\cite{ganformer} and segment the images using a pretrained segmentation model~\cite{hrnet}. 
Note that we cannot measure the mIoU for this baseline since the semantic layouts are predicted using the same pretrained network.

\begin{figure*}[t]
    \centering
    \includegraphics[width=\textwidth]{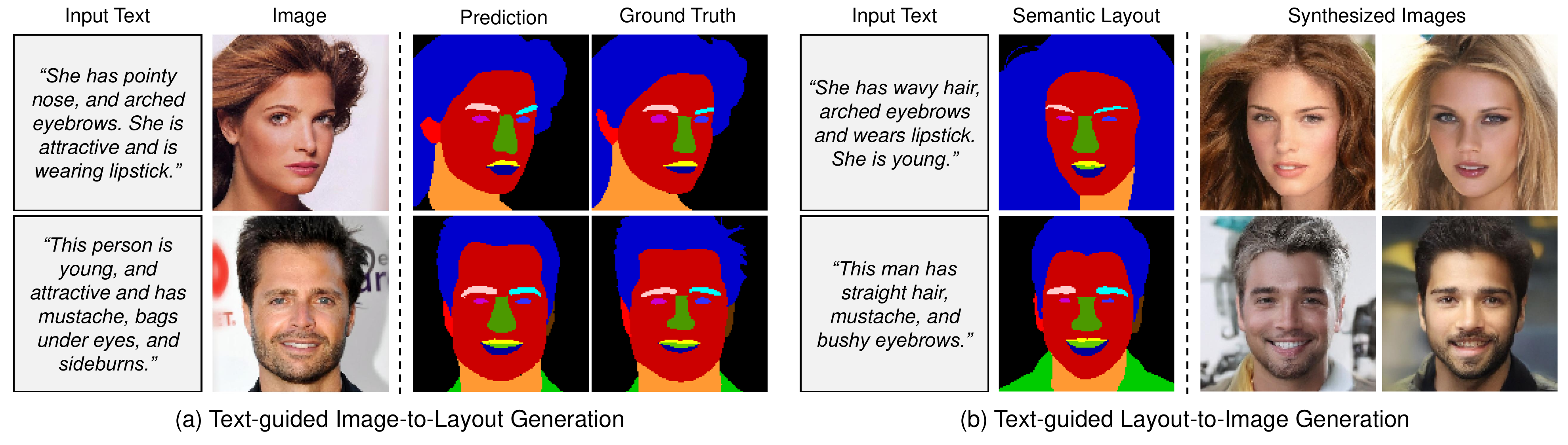}
    \caption{Cross-modal outpainting for (a) text-guided image-to-layout generation and (b) text-guided layout-to-image generation. Segmentation layouts are generated with $n=1$ resampling steps and images are generated with $n=5$ resampling steps for each timestep. }
    \label{fig:exp_outpainting}
    \vspace{-0.2cm}
\end{figure*}

As shown in \Cref{tab:exp_miou}, the Gaussian-categorical diffusion process is highly effective in modeling the joint distribution of images and layouts even for complex urban scenes. 
By using a unified diffusion process, we are able to generate image-layout pairs that exhibit high alignment, closely resembling the real distribution. 
The ability of the Gaussian-categorical diffusion to effectively model the joint distribution of images and layouts offers promising avenues for future research in generative modeling.
By leveraging the theoretical foundations established by our method, researchers can explore new approaches for dataset generation in a range of domains, from images and audios to semantic layouts and texts.

\subsection{Cross-modal outpainting}
RePaint~\cite{repaint} enables existing diffusion models to inpaint a masked image by iteratively denoising the masked region given the known image (\ie, condition image). 
Specifically, for each timestep $t$, images are inpainted as follows:
\begin{gather*}
    x_{t-1}^{\text{known}} \sim \gN(\sqrt{\bar{\alpha}_t}x_0, (1-\bar{\alpha}_t)\mI),\\
    x_{t-1}^{\text{unknown}} \sim \gN(\vmu_\theta(x_t, t), \mSigma_\theta(x_t, t)),\\
    x_{t-1} = m \odot x_{t-1}^{\text{known}} + (1-m) \odot x_{t-1}^{\text{unknown}},
\end{gather*}
where $m$ is the mask for the known image. 
To ensure consistency between the inpainted regions and known regions, Repaint iterates the denoising process $n$ times for each timestep. 

The Repaint technique allows us to use the Gaussian-categorical diffusion model as a text-guided layout-to-image generation model (\ie, semantic image synthesis) by considering the layouts as an image-layout pair with the image part masked. 
Similarly, we can perform text-guided image-to-layout generation (\ie, semantic segmentation) by masking the layout in the image-layout pair. 
As shown in \Cref{fig:exp_outpainting}, the Gaussian-categorical diffusion generates realistic images or layouts conditioned on text descriptions. 
The results demonstrate that a well-trained Gaussian-categorical diffusion can serve as a generative prior for conditional generation tasks. 
We describe the algorithm for cross-modal outpainting in the supplementary material. 

\section{Limitation}
Although the Gaussian-categorical diffusion offers means for achieving high text-image correspondence without training on web-scale text-image pairs, training a Gaussian-categorical diffusion model requires additional semantic layout annotations of images. 
However, with the assistance of recent data annotation tools~\cite{chen2022focalclick, sofiiuk2022reviving}, annotating existing data can be a cost-effective option for text-to-image generation in scenarios where obtaining web-scale text-image pairs is costly (\eg, medical images, urban scenes, and aerial images).

We observe that training the Gaussian-categorical diffusion model on the MS-COCO dataset~\cite{cocostuff} produces poor quality images and layouts. 
We suspect that this is due to the highly diverse scenes in the COCO dataset, with 171 categories in the semantic layouts. 
Analyzing the challenges of training on the MS-COCO dataset is a potential area for future research. 
Nevertheless, we propose an effective approach for text-to-image generation in data-scarce scenarios, where collecting data is expensive and annotating existing images is affordable.

\section{Conclusion}
In this paper, we define the Gaussian-categorical diffusion process to model the joint distribution of image-layout pairs. 
Our experiments demonstrate that the proposed model can ensure high text-image correspondence for text-to-image generation in specific domains, without relying on web-scale text-image pairs.
Our approach outperforms existing approaches in terms of image quality and text-image correspondence. 

Our visualizations of the internal representations of the Gaussian-categorical distribution demonstrate that the proposed model is aware of the semantics of the image, bridging the gap between highly semantic text descriptions and image pixels. 
Additionally, the high image-layout alignment of generated image-layout pairs and the results of cross-modal outpainting show that the model precisely captures the relationship between images and labels.

Overall, the Gaussian-categorical diffusion enables text-to-image models to achieve high text-image correspondence by leveraging semantic labels when trained on a specific domain with limited text-image pairs. 
Our proposed model can also be utilized as a generative prior for conditional generation tasks, such as text-guided semantic image synthesis and text-guided semantic segmentation. 

\section*{Acknowledgement}
This work was supported by the National Research Foundation of Korea (NRF) grant funded by the Korea government (MSIT) (No. NRF-2022R1A2B5B02001913), Electronics and Telecommunications Research Institute (ETRI) grant funded by the Korean government [22ZS1200, Fundamental Technology Research for Human-Centric Autonomous Intelligent Systems], and the KAIST-NAVER hypercreative AI center.

{\small
\bibliographystyle{ieee_fullname}
\bibliography{egbib}
}


\onecolumn
\fontsize{12}{15}\selectfont

\renewcommand\thesection{A.\arabic{section}} \setcounter{section}{0}
\renewcommand\thefigure{A.\arabic{figure}} \setcounter{figure}{0}
\renewcommand\thetable{A.\arabic{table}} \setcounter{table}{0}

\noindent{\Huge \textbf{Appendix}}
\vspace{2cm}


\section*{{\LARGE Table of Contents}}
\vspace{+1.0cm}
\begin{enumerate}[font=\Large]
    \item[]{\hypersetup{hidelinks}\nameref{sec:proofs}}\vspace{+0.8cm}
    \item[]{\hypersetup{hidelinks}\nameref{sec:noise}}\vspace{+0.8cm}
    \item[]{\hypersetup{hidelinks}\nameref{sec:sd-finetune}}\vspace{+0.8cm}
    \item[]{\hypersetup{hidelinks}\nameref{sec:clip}}\vspace{+0.8cm}
    \item[]{\hypersetup{hidelinks}\nameref{sec:sere}}\vspace{+0.8cm}
    \item[]{\hypersetup{hidelinks}\nameref{sec:cross-modal}}\vspace{+0.8cm}
    \item[]{\hypersetup{hidelinks}\nameref{sec:ablation}}\vspace{+0.8cm}
    \item[]{\hypersetup{hidelinks}\nameref{sec:quali}}\vspace{+0.8cm}
\end{enumerate}

\newpage

\section*{\LARGE A.1. Derivation of the Gaussian-categorical diffusion process}
\label{sec:proofs}
\vspace{+0.2cm}

In the following section, we provide detailed explanation of diffusion models including the categorical diffusion and the Gaussian-categorical diffusion.

\vspace{0.3cm}
\noindent{\textbf{\Large A.1.1. Categorical diffusion process}}
\label{sec:c-proof}
\vspace{0.2cm}

In this section, our final goal is to derive the posterior $q \cond{\rvy_{t-1}}{\rvy_t, \rvy_0}$ of the categorical diffusion, given the forward noising process. 
The forward process of the categorical diffusion process is defined as follows:
\begin{gather}
    \forall t \in [1,2, \ldots T], \quad \alpha_t \coloneqq 1 - \beta_t,
    \\
    q \cond{\rvy_t}{\rvy_{t-1}} \coloneqq \gC (\rvy_t; (1 - \beta_t) \rvy_{t-1} + \beta_t / K),\label{eq:c-def}
    \\
    \rvy_t \in \brkmid {1, 2, ..., K}^M \subset \R^{M}, \quad \mathbbm{1}[\rvy_t] \in \R^{M\times K},
\end{gather}
where $\beta_t$ is the noise schedule for each timestep, $K$ is the number of categories in the categorical distribution, and $M$ is the number of categorical variables. 
$\mathbbm{1}[\rvy_t]$ is the one-hot form of $\rvy_t$.

We will first prove $q \cond{\rvy_t}{\rvy_0} = \gC (\rvy_t ; \bar{\alpha}_t \rvy_0 + ( 1 - \bar{\alpha}_t) / K)$ through mathematical induction. 
The base case $t=1$ is evident though \Cref{eq:c-def} and let us assume the inductive case for $t-1$ where
\begin{gather}
    q \cond{\rvy_{t-1}}{\rvy_0} \coloneqq \gC (\rvy_{t-1} ; \bar{\alpha}_{t-1} \mathbbm{1}[\rvy_0] + ( 1 - \bar{\alpha}_{t-1}) / K) \quad \text{ where } \bar{\alpha}_t \coloneqq \prod_{s=1}^t \alpha_s.
\end{gather}

Then we can derive $q \cond{\rvy_t}{\rvy_0}$ as follows:
\begin{align}
    q \cond{\rvy_t}{\rvy_0} &= \sum_{\rvy_{t-1}} q \cond{\rvy_t}{\rvy_{t-1}, \rvy_0} q \cond{\rvy_{t-1}}{\rvy_0} 
    \\
    &= \sum_{\rvy_{t-1}} q \cond{\rvy_t}{\rvy_{t-1}} q \cond{\rvy_{t-1}}{\rvy_0} 
    \\
    &= \sum_{\rvy_{t-1}} \brkbig{\alpha_{t} \mathbbm{1}[\rvy_{t-1}] + ( 1 - \alpha_{t}) / K}_{\rvy_t} 
    \brkbig{\bar{\alpha}_{t-1} \mathbbm{1}[\rvy_0] + ( 1 - \bar{\alpha}_{t-1}) / K}_{\rvy_{t-1}} 
    \\
    &= \sum_{\rvy_{t-1}} \brkbig{\alpha_{t} \mathbbm{1}[\rvy_{t}] + ( 1 - \alpha_{t}) / K}_{\rvy_{t-1}} 
    \brkbig{\bar{\alpha}_{t-1} \mathbbm{1}[\rvy_0] + ( 1 - \bar{\alpha}_{t-1}) / K}_{\rvy_{t-1}}.
\end{align}
where $[\mTheta]_{\rvy_t}$ denotes the probability of event $\rvy_t$ in the categorical distribution parameterized with $\mTheta$.
By rewriting the summation as an inner product, we obtain
\begin{align}
    q \cond{\rvy_t}{\rvy_0} &= \brkbig{\alpha_{t} \mathbbm{1}[\rvy_{t}] + ( 1 - \alpha_{t}) / K} \cdot
    \brkbig{\bar{\alpha}_{t-1} \mathbbm{1}[\rvy_0] + ( 1 - \bar{\alpha}_{t-1}) / K}
    \\
    &= \bar{\alpha}_t \mathbbm{1}[\rvy_t] \cdot \mathbbm{1}[\rvy_0] + (1 - \alpha_t) \bar{\alpha}_{t-1} / K + (1 - \bar{\alpha}_{t-1}) \alpha_{t-1} / K + (1 - \alpha_t) (1 - \bar{\alpha}_{t-1}) / K
    \\
    &= \bar{\alpha}_t \mathbbm{1}[\rvy_t] \cdot \mathbbm{1}[\rvy_0] + (1 - \bar{\alpha}_t) / K
    \\
    &= \gC (\rvy_t; \bar{\alpha}_t \mathbbm{1}[\rvy_0] + (1 - \bar{\alpha}_t) / K),
\end{align}
which is the $t$ case of \Cref{eq:c-def}.
Through mathematical induction, we can conclude that $q \cond{\rvy_t}{\rvy_0} = \gC (\rvy_t ; \bar{\alpha}_t \mathbbm{1}[\rvy_0] + ( 1 - \bar{\alpha}_t) / K)$.

Next, we will derive the posterior $q \cond{\rvy_{t-1}}{\rvy_t, \rvy_0}$ using Bayes theorem as follows:
\begin{align}
    q \cond{\rvy_{t-1}}{\rvy_t, \rvy_0} &= \frac{q \cond{\rvy_{t}}{\rvy_{t-1}, \rvy_0} q \cond{\rvy_{t-1}}{\rvy_0}}{q \cond{\rvy_{t}}{\rvy_0}}
    \\
    &= \frac{q \cond{\rvy_{t}}{\rvy_{t-1}} q \cond{\rvy_{t-1}}{\rvy_0}}{q \cond{\rvy_{t}}{\rvy_0}}
    \\
    &= Z \brkbig{\alpha_t \mathbbm{1}[\rvy_{t-1}] + (1 - \alpha_t) / K}_{\rvy_t} \brkbig{\bar{\alpha}_{t-1} \mathbbm{1}[\rvy_0] + ( 1 - \bar{\alpha}_{t-1}) / K}_{\rvy_{t-1}} 
    \\
    &= Z \brkbig{\alpha_t \mathbbm{1}[\rvy_{t}] + (1 - \alpha_t) / K}_{\rvy_{t-1}} \brkbig{\bar{\alpha}_{t-1} \mathbbm{1}[\rvy_0] + ( 1 - \bar{\alpha}_{t-1}) / K}_{\rvy_{t-1}}
    \\
    &= \gC \bigl( \rvy_{t-1}; Z \brkbig{\alpha_t \mathbbm{1}[\rvy_{t}] + (1 - \alpha_t) / K} \odot \brkbig{\bar{\alpha}_{t-1} \mathbbm{1}[\rvy_0] + ( 1 - \bar{\alpha}_{t-1}) / K} \bigr).
\end{align}
Thus, the posterior $q \cond{\rvy_{t-1}}{\rvy_t, \rvy_0}$ is summarized as
\begin{gather}
    q \cond{\rvy_{t-1}}{\rvy_t, \rvy_0} = \gC (\rvy_{t-1} ; \widetilde{\mTheta}_t) \\
    \widetilde{\mTheta}_t \coloneqq Z [\alpha^\gCsmall_t \mathbbm{1}[\rvy_t] + (1 - \alpha^\gCsmall_t)/K] \odot [\bar{\alpha}^\gCsmall_t \mathbbm{1}[\rvy_0] + (1 - \bar{\alpha}^\gCsmall_{t-1}) / K].
\end{gather}

\vspace{0.3cm}
\noindent{\textbf{\Large A.1.2. Gaussian-categorical diffusion process}}
\label{sec:gc-proof}
\vspace{0.2cm}

We will derive the posterior $q \cond{\rvz_{t-1}}{\rvz_t, \rvz_0}$ of the Gaussian-categorical distribution, where the Gaussian distribution defined as follows:

\begin{gather}
\label{eq:supp_gc_dims}
    X, Y \sim \gN\gC \brk{\rvx, \rvy; \vmu, \mSigma, \mTheta},\\
    X=\brkbig{X_1, X_2, ... , X_N}\in\R^N,\nonumber\\
    Y=\brkbig{\emY_1, \emY_2, ... , \emY_M}\in\brkmid{1, 2, ..., K}^M,\nonumber\\        
    \vmu \in \R^{S \times N}, \mSigma \in \R^{S\times N\times N}, \mTheta \in \R^{M\times K}, \text{ and } S = K^M. \nonumber
\end{gather}

\begin{align}
    \gN\gC(\rvx, \rvy; \vmu, \mSigma, \mTheta) = \brk{ \prod_{i=1}^M \mTheta_{\scaleto{i}{6pt}\,,\, \scaleto{\rvy}{6pt}_{\scaleto{i}{4.5pt}}} } \brk{2\pi}^{-\frac{N}{2}} \abs{\mSigma_\rvy}^{-\frac{1}{2}} \exp{ \Big( -\frac{1}{2} (\rvx - \vmu_\rvy)^\top \mSigma^{-1}_\rvy \brk{\rvx - \vmu_\rvy} \Big) }.
\end{align}
and the forward noising process for the Gaussian-categorical diffusion is defined as
\begin{gather}
    \forall t\in[1,2,\ldots, T], \quad \alpha_t^\gNsmall \coloneqq 1 - \beta_t^\gNsmall, \quad \alpha_t^\gCsmall \coloneqq 1 - \beta_t^\gCsmall, \quad \text{ and } \quad \rvz_t \coloneqq (\rvx_t, \rvy_t),
    \\
    q \cond{\rvz_t}{\rvz_{t-1}} \coloneqq \gN\gC \brk{\rvz_t; \bigl[ \sqrt{1 - \beta_t^{\gNsmall}} \rvx_{t-1} \bigr]_{\times S}, \bigl[ \beta_{t}^\gNsmall \mI \bigr]_{\times S}, (1 - \beta_t^\gCsmall)\mathbbm{1}[\rvy_{t-1}] + \beta_t^\gCsmall / K }.\label{eq:gc-def}
\end{gather}

We will first prove that $q \cond{\rvz_t}{\rvz_0} = \gN\gC \Bigl( \rvz_{t} ; \bigl[ \sqrt{ \bar{\alpha}^\gNsmall_t } \rvx_{0} \bigr]_{\times S}, \bigl[ (1 - \bar{\alpha}^\gNsmall_t) \mI \bigr]_{\times S}, (1 - \bar{\alpha}^\gCsmall_t) \mathbbm{1}[\rvy_{0}] + \bar{\alpha}^\gCsmall_t / K \Bigl)$  where $\bar{\alpha}_t \coloneqq \prod_{s=1}^t \alpha_s$. 
We will prove this using mathematical induction, where the base case $t=1$ is defined in \Cref{eq:gc-def}.
Let us assume the inductive case for $t-1$,
\begin{gather}
    q \cond{\rvz_{t-1}}{\rvz_0} = \gN\gC \Bigl( \rvz_{t-1} ; \bigl[ \sqrt{ \bar{\alpha}^\gNsmall_{t-1} } \rvx_{0} \bigr]_{\times S}, \bigl[ (1 - \bar{\alpha}^\gNsmall_{t-1}) \mI \bigr]_{\times S}, (1 - \bar{\alpha}^\gCsmall_{t-1}) \mathbbm{1}[\rvy_{0}] + \bar{\alpha}^\gCsmall_{t-1} / K \Bigl).
\end{gather}
Then we can derive $q \cond{\rvz_{t}}{\rvz_0}$ as follows:
\begin{align}
    q &\cond{\rvz_{t}}{\rvz_0} 
    \\
    &= \int q \cond{\rvz_{t}}{\rvz_{t-1}, \rvz_0} q \cond{\rvz_{t-1}}{\rvz_0} d\rvz_{t-1}
    \\
    &= \int q \cond{\rvz_{t}}{\rvz_{t-1}} q \cond{\rvz_{t-1}}{\rvz_0} d\rvz_{t-1}
    \\
    &= \sum_{\rvy_{t-1}} \int \gN\gC(\rvz_{t}; \bigl[ \vmu_{t|t-1} \bigr]_{\times S}, \bigl[ \mSigma_{t|t-1} \bigr]_{\times S}, \mTheta_{t|t-1}) \cdot \gN\gC(\rvz_{t-1}; \bigl[ \vmu_{t-1|0} \bigr]_{\times S}, \bigl[ \mSigma_{t-1|0} \bigr]_{\times S}, \mTheta_{t-1|0}) d\rvx_{t-1},
\end{align}
where $\mTheta_{i|j} \coloneqq (1 - \beta^\gCsmall_i) \mathbbm{1}[\rvy_{j}] + \beta^\gCsmall_i / K$, and $[ \rvv ]_{\times S}$ indicates row-wise duplication of a vector $\rvv$ (\ie, $[ \rvv, \rvv, ..., \rvv]^T$). By decomposing the Gaussian-categorical into a Gaussian distribution and a categorical distribution, we can write the equation as follows:
\begin{align}
    q &\cond{\rvz_{t}}{\rvz_0} 
    \\
    &= \sum_{\rvy_{t-1}} \int \Bigl( \gC(\rvy_{t}; \mTheta_{t|t-1}) \cdot \gN(\rvx_{t}; \vmu_{t|t-1}, \mSigma_{t|t-1}) \Bigr) \cdot \Bigl( \gC(\rvy_{t-1}; \mTheta_{t-1|0}) \cdot \gN(\rvx_{t-1}; \vmu_{t-1|0}, \mSigma_{t-1|0}) \Bigr) d\rvx_{t-1}
    \\
    &= \sum_{\rvy_{t-1}} \gC(\rvy_{t}; \mTheta_{t|t-1}) \cdot \gC(\rvy_{t-1}; \mTheta_{t-1|0})
    \int \gN(\rvx_{t}; \vmu_{t|t-1}, \mSigma_{t|t-1}) \cdot \gN(\rvx_{t-1}; \vmu_{t-1|0}, \mSigma_{t-1|0}) d\rvx_{t-1}
    \\
    &= \gC (\rvy_t; \bar{\alpha}_t^{\gCsmall} \mathbbm{1}[\rvy_0] + (1 - \bar{\alpha}_t^{\gCsmall}) / K) \cdot \gN (\rvx_t; \sqrt{\bar{\alpha}_t^{\gNsmall}} \rvx_0, (1 - \bar{\alpha}_t^{\gNsmall}) \mI )
    \\
    &= \gN\gC \Bigl( \rvz_{t} ; \bigl[ \sqrt{ \bar{\alpha}^\gNsmall_t } \rvx_{0} \bigr]_{\times S}, \bigl[ (1 - \bar{\alpha}^\gNsmall_t) \mI \bigr]_{\times S}, (1 - \bar{\alpha}^\gCsmall_t) \mathbbm{1}[\rvy_{0}] + \bar{\alpha}^\gCsmall_t / K \Bigl),
\end{align}
where $\vmu_{i|j} \coloneqq \sqrt{1 - \beta^\gNsmall_i} \rvx_{j}$ and $\mSigma_{i|j} \coloneqq \beta^\gNsmall_i \mI$. 
Through mathematical induction, we can conclude that $q \cond{\rvz_t}{\rvz_0} = \gN\gC \Bigl( \rvz_{t} ; \bigl[ \sqrt{ \bar{\alpha}^\gNsmall_t } \rvx_{0} \bigr]_{\times S}, \bigl[ (1 - \bar{\alpha}^\gNsmall_t) \mI \bigr]_{\times S}, (1 - \bar{\alpha}^\gCsmall_t) \mathbbm{1}[\rvy_{0}] + \bar{\alpha}^\gCsmall_t / K \Bigl)$.

\vspace{+0.2cm}
Next, we will derive the posterior $q \cond{\rvz_{t-1}}{\rvz_t, \rvz_0}$ using Bayes theorem,
\begin{align}
    q \cond{\rvz_{t-1}}{\rvz_t, \rvz_0} &= \frac{q \cond{\rvz_{t}}{\rvz_{t-1}, \rvz_0} q \cond{\rvz_{t-1}}{\rvz_0}}{q \cond{\rvz_{t}}{\rvz_0}}
    \\
    &= \frac{q \cond{\rvz_{t}}{\rvz_{t-1}} q \cond{\rvz_{t-1}}{\rvz_0}}{q \cond{\rvz_{t}}{\rvz_0}}
    \\
    &= \frac{\gN\gC(\rvz_{t}; \bigl[ \vmu_{t|t-1} \bigr]_{\times S}, \bigl[ \mSigma_{t|t-1} \bigr]_{\times S}, \mTheta_{t|t-1}) \cdot \gN\gC(\rvz_{t-1}; \bigl[ \vmu_{t-1|0} \bigr]_{\times S}, \bigl[ \mSigma_{t-1|0} \bigr]_{\times S}, \mTheta_{t-1|0})}{\gN\gC(\rvz_{t}; \bigl[ \vmu_{t|0} \bigr]_{\times S}, \bigl[ \mSigma_{t|0} \bigr]_{\times S}, \mTheta_{t|0})}.
\end{align}
We again decompose the Gaussian-categorical diffusion into a Gaussian distribution and a categorical distribution
\begin{align}
    q &\cond{\rvz_{t-1}}{\rvz_t, \rvz_0}
    \\
    &= \frac{\Bigl( \gC(\rvy_{t}; \mTheta_{t|t-1}) \cdot \gN(\rvx_{t}; \vmu_{t|t-1}, \mSigma_{t|t-1}) \Bigr) \cdot \Bigl( \gC(\rvy_{t-1}; \mTheta_{t-1|0}) \cdot \gN(\rvx_{t-1}; \vmu_{t-1|0}, \mSigma_{t-1|0}) \Bigr)}{\gC(\rvy_{t}; \mTheta_{t|0}) \cdot \gN(\rvx_{t}; \vmu_{t|0}, \mSigma_{t|0})}
    \\
    &= \frac{\gC(\rvy_{t}; \mTheta_{t|t-1}) \cdot \gC(\rvy_{t-1}; \mTheta_{t-1|0})}{\gC(\rvy_{t}; \mTheta_{t|0})}
    \cdot \frac{\gN(\rvx_{t}; \vmu_{t|t-1}, \mSigma_{t|t-1}) \cdot \gN(\rvx_{t-1}; \vmu_{t-1|0}, \mSigma_{t-1|0})}{\gN(\rvx_{t}; \vmu_{t|0}, \mSigma_{t|0})}
    \\
    &= \gC(\rvy_{t-1}; \widetilde{\mTheta}_t) \cdot \gN(\rvx_{t-1}; \widetilde{\vmu}_t, \widetilde{\mSigma}_t)
    \\
    &= \gN\gC (\rvz_{t-1}; \bigl[\widetilde{\vmu}_t\bigr]_{\times S}, \bigl[\widetilde{\mSigma}_t\bigr]_{\times S}, \widetilde{\mTheta}_t),
\end{align}
\begin{gather}
    \widetilde{\vmu}_t \coloneqq \frac{\sqrt{\bar{\alpha}^{\gNsmall}_{t-1}}\beta_t^{\gNsmall}}{1 - \bar{\alpha}^{\gNsmall}_t} \rvx_0 + \frac{\sqrt{\alpha^{\gNsmall}_t}(1-\bar{\alpha}^{\gNsmall}_{t-1})}{1 - \bar{\alpha}^{\gNsmall}_t} \rvx_t,
    \\
    \widetilde{\mSigma}_t \coloneqq \bigl( (1 - \bar{\alpha}^\gNsmall_{t-1}) \beta^\gNsmall_t / ( 1 - \bar{\alpha}^\gNsmall_{t}) \bigr) \mI,
    \\
    \widetilde{\mTheta}_t \coloneqq Z [\alpha^\gCsmall_t \mathbbm{1}[\rvy_t] + (1 - \alpha^\gCsmall_t)/K] \odot [\bar{\alpha}^\gCsmall_t \mathbbm{1}[\rvy_0] + (1 - \bar{\alpha}^\gCsmall_{t-1}) / K],
\end{gather}

The posterior distribution $q \cond{\rvz_{t-1}}{\rvz_t, \rvz_0}$ can be summarized as follows:
\begin{align}
    q \cond{\rvz_{t-1}}{\rvz_t, \rvz_0} = \gN\gC \Bigl( \rvz_{t-1}; \bigl[\widetilde{\vmu}_t \bigr]_{\times S}, \bigl[ \widetilde{\mSigma}_t \bigr]_{\times S}, \widetilde{\mTheta}_t \Bigr), \hspace{-1pt}
\end{align}
where $Z$ is a normalizing constant. 
We approximate the reverse process by matching $\widetilde{\vmu}_\theta(\rvz_t)$, $\widetilde{\mSigma}_\theta(\rvz_t)$, and $\mTheta_\theta(\rvz_t)$. 

Finally, minimizing the KL divergence term $D_{KL} \bigl( q\cond{\rvz_{t-1}}{\rvz_t, \rvz_0} \parallel p_\theta \cond{\rvz_{t-1}}{\rvz_t} \bigr)$ can be decomposed into two separate terms for the Gaussian variable and the categorical variable as follows:
\begin{align}
    &D_{KL} \bigl( q\cond{\rvz_{t-1}}{\rvz_t, \rvz_0} \parallel p_\theta \cond{\rvz_{t-1}}{\rvz_t} \bigr)\\
    &= \int q \cond{\rvz_{t-1}}{\rvz_t, \rvz_0} \log {\frac{q \cond{\rvz_{t-1}}{\rvz_t, \rvz_0}}{p_\theta \cond{\rvz_{t-1}}{\rvz_t}}} d\rvz_{t-1} \\
    &= \int \gN\gC (\rvz_{t-1}; \bigl[\widetilde{\vmu}_t\bigr]_{\times S}, \bigl[\widetilde{\mSigma}_t\bigr]_{\times S}, \widetilde{\mTheta}_t) \log{\frac{\gN\gC (\rvz_{t-1}; \bigl[\widetilde{\vmu}_t\bigr]_{\times S}, \bigl[\widetilde{\mSigma}_t\bigr]_{\times S}, \widetilde{\mTheta}_t)}{\gN\gC (\rvz_{t-1}; \bigl[\widetilde{\vmu}_\theta (\rvz_t) \bigr]_{\times S}, \bigl[\widetilde{\mSigma}_\theta (\rvz_t) \bigr]_{\times S}, \mTheta_\theta (\rvz_t))}} d\rvz_{t-1}\\
    &= \int \gC(\rvy_{t-1}; \widetilde{\mTheta}_t) \cdot \gN(\rvx_{t-1}; \widetilde{\vmu}_t, \widetilde{\mSigma}_t) 
    \log {\frac{\gC(\rvy_{t-1}; \widetilde{\mTheta}_t) \cdot \gN(\rvx_{t-1}; \widetilde{\vmu}_t, \widetilde{\mSigma}_t)}
    {\gC(\rvy_{t-1}; \mTheta_\theta (\rvz_t)) \cdot \gN(\rvx_{t-1}; \widetilde{\vmu}_\theta (\rvz_t), \widetilde{\mSigma}_\theta (\rvz_t))}}
    d\rvz_{t-1}\\
    &= \int \gC(\rvy_{t-1}; \widetilde{\mTheta}_t) \cdot 
    \gN(\rvx_{t-1}; \widetilde{\vmu}_t, \widetilde{\mSigma}_t) 
    \log {\frac{\gC(\rvy_{t-1}; \widetilde{\mTheta}_t)}{\gC(\rvy_{t-1}; \mTheta_\theta (\rvz_t))}} d\rvz_{t-1} \nonumber\\
    &\quad + \int \gC(\rvy_{t-1}; \widetilde{\mTheta}_t) \cdot \gN(\rvx_{t-1}; \widetilde{\vmu}_t, \widetilde{\mSigma}_t) 
    \log {\frac{\gN(\rvx_{t-1}; \widetilde{\vmu}_t, \widetilde{\mSigma}_t)}
    {\gN(\rvx_{t-1}; \widetilde{\vmu}_\theta (\rvz_t), \widetilde{\mSigma}_\theta (\rvz_t))}}
    d\rvz_{t-1}\\
    &= \int \gC(\rvy_{t-1}; \widetilde{\mTheta}_t) \cdot 
    \log {\frac{\gC(\rvy_{t-1}; \widetilde{\mTheta}_t)}{\gC(\rvy_{t-1}; \mTheta_\theta (\rvz_t))}} d\rvy_{t-1} \nonumber\\
    &\quad + \int \gN(\rvx_{t-1}; \widetilde{\vmu}_t, \widetilde{\mSigma}_t) 
    \log {\frac{\gN(\rvx_{t-1}; \widetilde{\vmu}_t, \widetilde{\mSigma}_t)}
    {\gN(\rvx_{t-1}; \widetilde{\vmu}_\theta (\rvz_t), \widetilde{\mSigma}_\theta (\rvz_t))}}
    d\rvx_{t-1}\\
    &= D_{KL}( \gC(\rvy_{t-1}; \widetilde{\mTheta}_t) \parallel \gC(\rvy_{t-1}; \mTheta_\theta (\rvz_t)) ) 
    + D_{KL} (\gN(\rvx_{t-1}; \widetilde{\vmu}_t, \widetilde{\mSigma}_t) \parallel \gN(\rvx_{t-1}; \widetilde{\vmu}_\theta (\rvz_t), \widetilde{\mSigma}_\theta (\rvz_t))) \\
    &= \E_q \Bigl[ \frac{1}{2\sigma_t^2} \norm{\widetilde{\vmu}_t - \widetilde{\vmu}_\theta (\rvz_t)}^2 \Bigr] + D_{KL} ( \widetilde{\mTheta}_t \parallel \mTheta_\theta(\rvz_t) ) + C
\end{align}

\begin{figure}[ht]
    \centering
    \includegraphics[width=\textwidth]{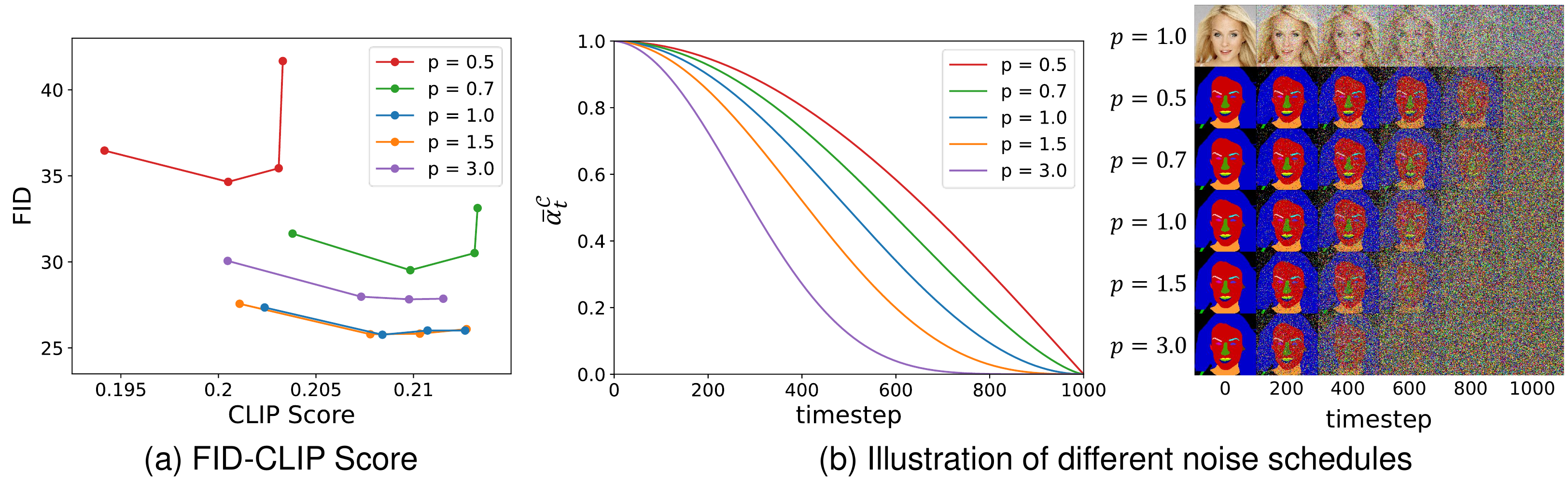}
    \caption{(a) FID-CLIP score pairs for different noise schedules $\beta^{\gCsmall}$. FID and CLIP scores are measured in $128\times 128$ resolution. (b) The illustration of different noise schedules. A larger $p$ indicates stronger noise near $t=1000$.}
    \label{fig:beta_search}
\end{figure}

\section*{\LARGE A.2. Noise schedules of the Gaussian-categorical diffusion process}
\label{sec:noise}
\vspace{0.2cm}

The Gaussian-categorical diffusion process can have different noise schedules for $\beta^{\gCsmall}$ and $\beta^\gNsmall$ as defined in \Cref{eq:gc-def}. 
In order to search for a reasonable noise schedule, we train the Gaussian-categorical diffusion model on different schedules for $\beta^\gCsmall$, relative to the Gaussian noise schedule $\beta^\gNsmall$. 
Specifically, we fix $\beta^\gNsmall$ as the cosine noise schedule~\cite{iddpm} and set $\beta^\gCsmall$ as a function of a $p^{\text{th}}$ power of $\beta^\gNsmall$, in other words $\beta^\gCsmall \coloneqq (\beta^\gNsmall)^p$, which are plotted in \Cref{fig:beta_search} (b). 
In \Cref{fig:beta_search}, we present the FID-CLIP score of these results at the $128\times 128$ resolution on the CelebA-HQ dataset~\cite{celeba}.
Overall, choosing $p$ near 1 is a good choice for achieving text-image correspondence. 
We leave further analysis on noise scheduling between different modalities as a future research topic.

\begin{figure}[ht]
    \centering
    \includegraphics[width=0.45\textwidth]{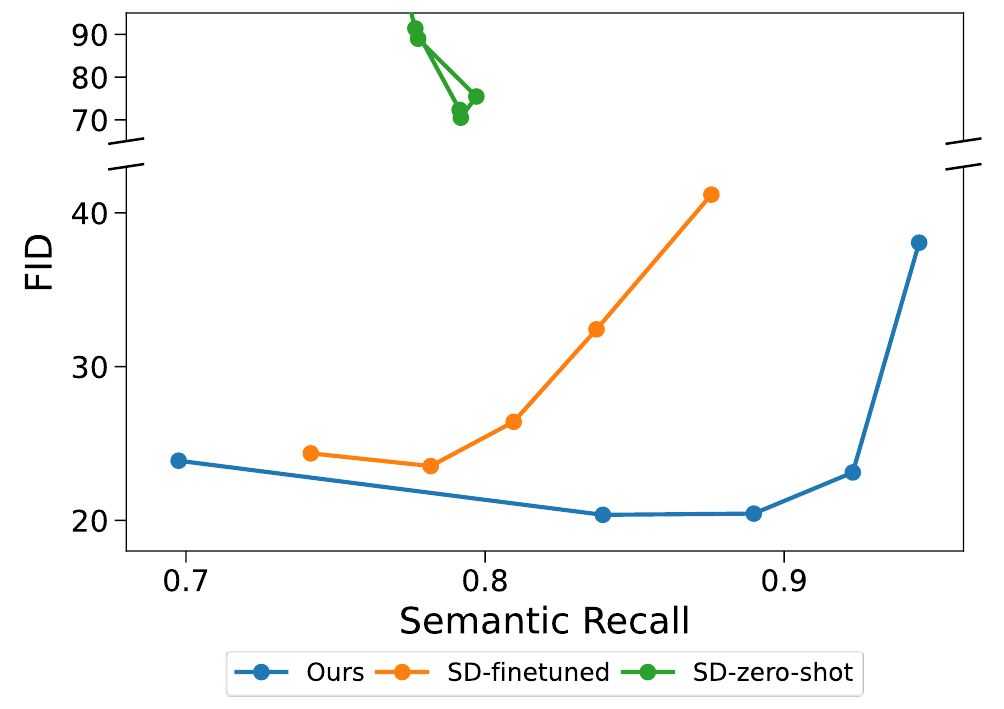}
    \caption{FID-Semantic Recall of the Gaussian-categorical diffusion model compared to the results generated by the Stable Diffusion model finetuned on Cityscapes~\cite{cityscapes} (SD-finetuned) and zero-shot text-to-image generation of the pretrained Stable Diffusion (SD-zero-shot). We use the Stable Diffusion v1.4 for both zero-shot generation and finetuning.}
    \label{fig:sd_finetune}
\end{figure}

\section*{\LARGE A.3. Comparison with Stable Diffusion}
\label{sec:sd-finetune}
\vspace{0.2cm}

Recently, finetuning a general-purpose text-to-image generation model using domain-specific datasets has shown great success in generating high-quality images with strong text-image correspondence. 
Specifically, the Stable Diffusion project provides a large pretrained Latent Diffusion Model (LDM)~\cite{ldm} trained on a web-scale dataset, the LAION 5B~\cite{laion}, that is capable of generating artistic images. 
In this section, we demonstrate the limitation of finetuning a generative model in cases of significant domain gaps. 
We finetune Stable Diffusion v1.4 using the Cityscapes dataset and report the FID-Semantic Recall pairs in \Cref{fig:sd_finetune}. 
We also provide zero-shot text-to-image generation results for comparison. 
While finetuning stable diffusion can be effective in natural domains such as the MM CelebA-HQ, it should not be viewed as an all-encompassing solution for addressing issues in text-to-image generation. 
Neither finetuning Stable Diffusion nor zero-shot text-to-image generation exhibits a low FID or a high Semantic Recall for generating the urban scenes of Cityscapes~\cite{cityscapes}.
Training a Gaussian-categorical diffusion model can be an effective approach for generating unique domains such as medical images or aerial photos.

\begin{figure}[ht]
    \centering
    \includegraphics[width=\textwidth]{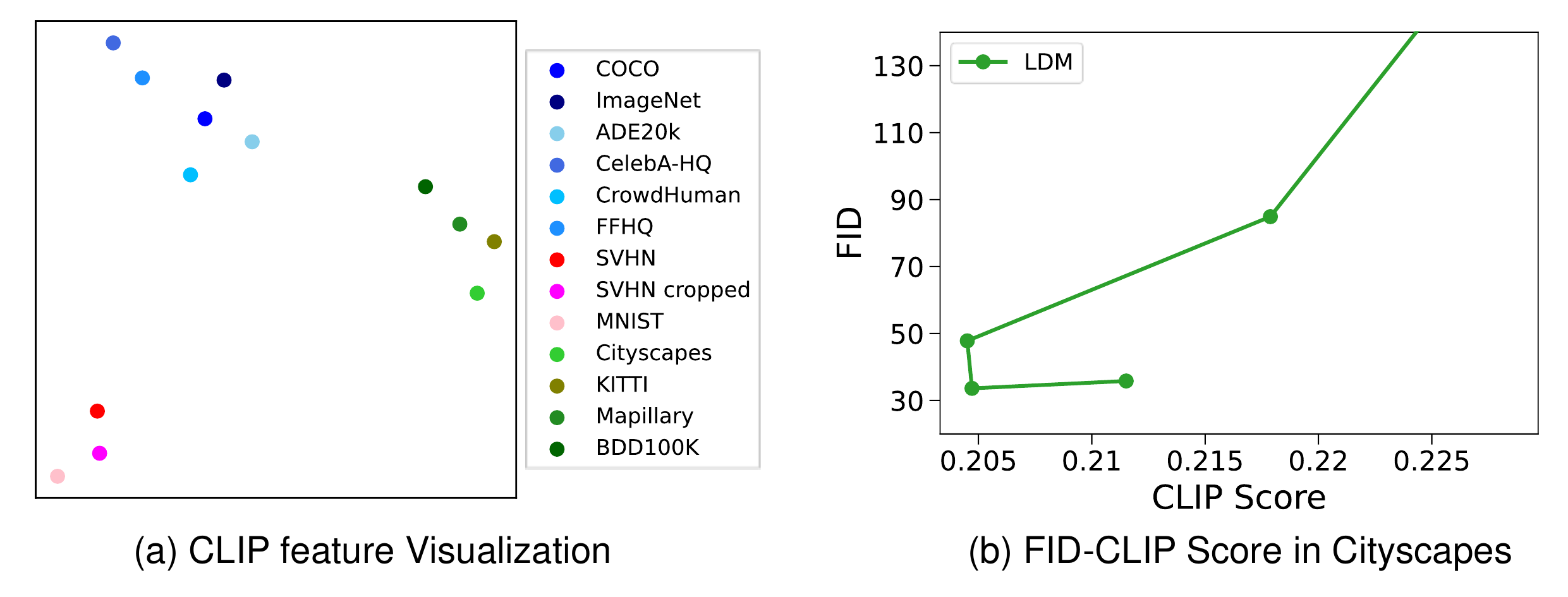}
    \caption{(a) Visualization of CLIP features from different datasets using t-SNE. While the CelebA-HQ dataset closely clusters with several large-scale image datasets such as the ImageNet and MS COCO dataset, urban scene datasets such as Cityscapes or BDD100K form distinct clusters. (b) CLIP scores display inconsistent trends when measured on the Cityscapes dataset.}
    \label{fig:clip_domain_gap}
\end{figure}

\section*{\LARGE A.4. Visualizing the domain gaps in CLIP scores}
\label{sec:clip}
\vspace{0.2cm}

The CLIP score~\cite{clipscore} is a reliable measure in most cases for evaluating the quality of text-to-image generation in natural domains such as the MS COCO~\cite{cocostuff}. 
However, in certain cases, the CLIP model~\cite{clip} may have poor generalization abilities for specific domains with significant differences from its training data. 
Since the train dataset of CLIP is not publicly available for this analysis, we replace it with the MS COCO dataset which contains diverse images of different scenes. 
As shown in \Cref{fig:clip_domain_gap} (a), we plot the features from the CLIP image encoder~\cite{clip} for different datasets using the t-SNE visualization technique~\cite{tsne}. 
Each point in \Cref{fig:clip_domain_gap} (a) represents the averaged CLIP features from a single dataset.
While general image datasets such as the ImageNet~\cite{imagenet}, ADE20K~\cite{ade}, and the CelebA-HQ~\cite{celeba} are closely clustered to the MS COCO dataset, other datasets such as the urban scene datasets (\eg, Cityscapes~\cite{cityscapes} and BDD100K~\cite{bdd100k}) or the number datasets (\eg, MNIST~\cite{mnist} and SVHN~\cite{svhn}) form distinct clusters apart from the MS COCO dataset~\cite{cocostuff}. 

This indicates that the Cityscapes dataset may have a domain gap significantly large enough to render the CLIP score unreliable. 
As shown in \Cref{fig:clip_domain_gap} (b), FID-CLIP score pairs for the Latent Diffusion Model (LDM)~\cite{ldm} display inconsistent trends of increase and decrease as the guidance scale increases. 
Thus, we do not use the CLIP score to evaluate the Cityscapes text-to-image generation and instead use the Semantic Recall. 

\begin{figure}[ht]
    \centering
    \includegraphics[width=\textwidth]{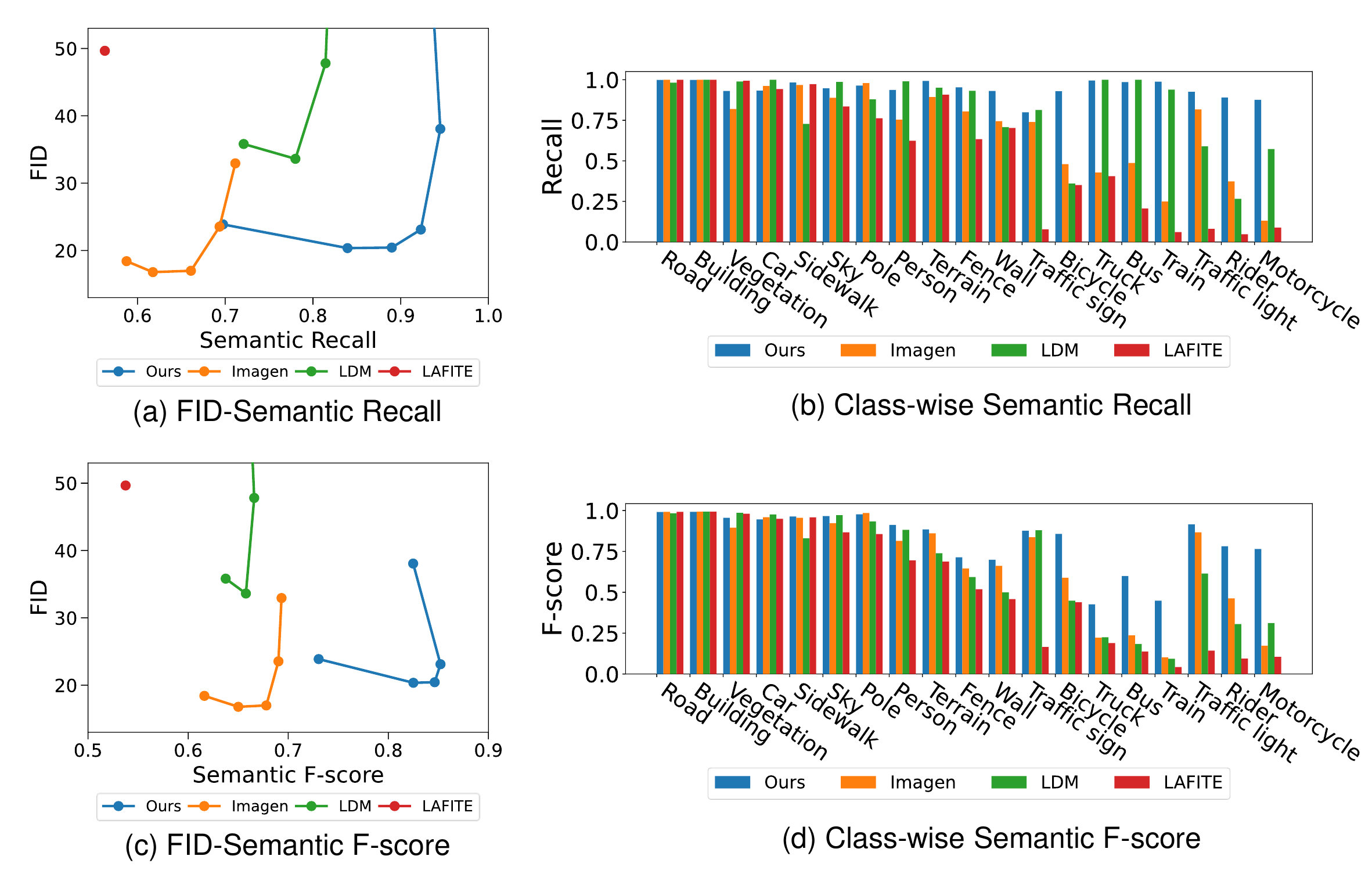}
    \caption{(a) FID-Semantic Recall for the Cityscapes dataset and (b) detailed class-wise Semantic Recall. (c) FID-Semantic F-score for the Cityscapes dataset and (c) detailed class-wise Semantic Recall. Classes are sorted from the most occurring classes (left) to the least occurring (right). The Gaussian-categorical diffusion model outperforms existing baselines by a large margin in the Semantic F-score, indicating that our approach does not overly generate objects. }
    \label{fig:semantic_f1}
\end{figure}

\section*{\LARGE A.5. Semantic Recall in Cityscapes}
\label{sec:sere}
\vspace{0.2cm}

To compensate for the limitations of the CLIP score when evaluating datasets with large domain gaps, we introduce the Semantic Recall which evaluates the generation of specific semantic categories specified in the test description. 
The Semantic Recall is the average ratio of correctly detected classes in the generated image to the total number of classes in the ground-truth layouts,
\begin{equation*}
    \text{Semantic Recall} \coloneqq \frac{1}{\mid \mathcal{G}\mid}\sum_{x_i, y_i \in \mathcal{G}}\frac{\mid \text{Classes in }F(x_i) \cap \text{ Classes in } y_i\mid}{\mid \text{Classes in }y_i\mid},
\end{equation*}
where $\mathcal{G}$ is the set of generated image-layout pairs $(x_i, y_i)$ and $\mid\hspace{-2pt}\cdot\hspace{-2pt}\mid$ indicates the cardinality of a given set. 
$F(\cdot)$ is the pretrained semantic segmentation model~\cite{hrnet}. 
We provide full details of the Semantic Recall for each class in \Cref{fig:semantic_f1} (b). 
The Gaussian-categorical diffusion model is especially effective for generating less frequently encountered classes such as the \emph{Motorcycle} and \emph{Traffic light} classes.

In this section, we also report the Semantic F-score as an evaluation measure for the semantic accuracy of the generated image. 
The Semantic F-score is similar to the Semantic Recall but uses the F-score, which takes both recall and precision into account as:
\begin{equation*}
    \text{Semantic F-score} \coloneqq \frac{2}{\text{Semantic Recall}^{-1} + \text{ Semantic Precision}^{-1}},
\end{equation*}
where Semantic Precision is calculated similarly to the Semantic Recall.
While the Semantic Recall is useful for detecting the existence of certain objects, it may overcompensate for verbose generation. 
For instance, a text-to-image generation model that generates all semantic classes regardless of the text condition may achieve a high recall without understanding the text description. 
Therefore, we use the F-score to evaluate whether a text-to-image generation model precisely generates the classes specified in the text description. 
The results in \Cref{fig:semantic_f1} (c) demonstrate that the Gaussian-categorical diffusion model outperforms existing text-to-image in the Cityscapes~\cite{cityscapes} dataset, exhibiting a high F-score and a low FID. 
This suggests that our model does not overly generate semantic classes regardless of the text description.

\section*{\LARGE A.6. Quantitative results for cross-modal outpainting}
\label{sec:cross-modal}
\vspace{0.2cm}
As demonstrated in the main paper, a well-trained Gaussian-categorical diffusion is capable of performing text-guided segmentation and layout-to-image generation. 
The key idea is to view an image or a layout as a masked image-layout pair and inpaint the masked modality using the RePaint technique~\cite{repaint}.
The detailed algorithm following RePaint~\cite{repaint} is provided in \Cref{alg:cross-modal}. 
We also compare the quantitative comparison of the results for segmentation and layout-to-image generation on the CelebA-HQ dataset~\cite{celeba, pggan} in \Cref{tab:segmentation} and \Cref{tab:sis}. 
We train a segmentation (\ie, Deeplab v3~\cite{deeplabv3}) and a layout-to-image generation model (\ie, OASIS~\cite{oasis}) on the MM CelebA-HQ-25. 
While the Gaussian-categorical diffusion does not outperform models dedicated to segmentation or layout-to-image generation, it yields reasonable quantitative results which suggest that the Gaussian-categorical diffusion can serve as a generative prior for tasks other than text-to-image generation. 
Additionally, we find that training the Gaussian-categorical diffusion with a lower $p$ value leans towards better layout-to-image generation while a higher $p$ value leads to better segmentation performance. 
In this manner, extreme values of $p$ (\ie, $p=0$ and $p\rightarrow\infty$) are equivalent to training a conditional generation model (\ie, layout-to-image and semantic segmentation).

\begin{algorithm}[ht]
\begin{algorithmic}[1]
\State $\rvz_T \sim \gN\gC(\rvx, \rvy; \textbf{0}, \mI, \mTheta)$
\State $t \gets T$
\While {$t > 0$}
\State $n \gets N$
\While {$n > 0$}
\vspace{+0.1cm}
\State $\rvz^{\text{known}}_{t-1} \sim \gN\gC(\rvz_{t-1}; \bigl[\vmu_{t-1|0}\bigr]_{\times S}, \bigl[\mSigma_{t-1|0}\bigr]_{\times S}, \mTheta_{t-1|0})$ \algorithmiccomment{Apply noise to known area $\rvz^{\text{known}}$}
\vspace{+0.2cm}
\State $\rvz^{\text{unknown}}_{t-1} \sim \gN\gC(\rvz_{t-1}; \bigr[\widetilde{\vmu}_\theta(\rvz_t)\bigl]_{\times S}, \bigr[\widetilde{\mSigma}_\theta(\rvz_t)\bigl]_{\times S}, \mTheta_\theta(\rvz_t))$\algorithmiccomment{Denoise single step $\rvz_t$}
\vspace{+0.2cm}
\State $\rvz_{t-1} = m \odot \rvz^{\text{known}}_{t-1} + (1-m) \odot \rvz^{\text{unknown}}_{t-1}$\algorithmiccomment{Update unknown area}
\vspace{+0.2cm}
\If{$n < N$ and $t > 1$}
    \vspace{+0.2cm}
    \State $\rvz_{t} \sim \gN\gC(\rvz_t; \bigr[\vmu_{t|t-1}\bigl]_{\times S}, \bigr[\mSigma_{t|t-1}\bigl]_{\times S}, \mTheta_{t|t-1})$\algorithmiccomment{Resample timestep $t$}
\EndIf

\State $n \gets n-1$
\EndWhile
\State $t \gets t-1$
\EndWhile
\end{algorithmic}
\caption{Cross-modal outpainting for conditional generation.}
\label{alg:cross-modal}
\end{algorithm}

\begin{table}[ht]
\parbox[t]{.45\linewidth}{
\centering
\begin{tabular}{@{}x{3.6cm}x{2.3cm}@{}}
\toprule
Method                     & mIoU $\uparrow$      \\ \midrule
Deeplab v3~\cite{deeplabv3}& 73.88     \\ \midrule
Ours $p=0.5$               & 32.52     \\
Ours $p=1.0$               & 51.56     \\
Ours $p=3.0$               & 59.82     \\ \bottomrule
\end{tabular}
\vspace{+0.2cm}
\caption{Quantitative results for semantic segmentation on the 25\% of the MM CelebAMask-HQ dataset~\cite{celebamask}. Segmentation predictions are generated by resampling noise 5 times for each timestep ($N=5$).}
\label{tab:segmentation}
}
\hfill
\parbox[t]{.45\linewidth}{
\centering
\begin{tabular}{@{}x{3.2cm}x{1.4cm}x{1.4cm}@{}}
\toprule
Method            & FID $\downarrow$ & mIoU $\uparrow$ \\ \midrule
OASIS~\cite{oasis}& 20.64            & 77.35\\  \midrule
Ours $p=0.5$      & 30.45            & 71.51 \\
Ours $p=1.0$      & 33.25            & 66.81 \\
Ours $p=3.0$      & 47.89            & 40.09 \\ \bottomrule
\end{tabular}
\vspace{+0.2cm}
\caption{Quantitative results for layout-to-image generation on MM CelebAMask-HQ-25 dataset~\cite{celebamask}. mIoU is measured between the input layout and the segmentation results of the generated image using a pretrained HRNet~\cite{hrnet}.}
\label{tab:sis}
}
\end{table}

\section*{\LARGE A.7. Ablation study and additional baselines}
\label{sec:ablation}
\vspace{0.2cm}

In this section, we provide results for different text-to-image generation approaches and compare them against our approach. 
First, we train a Gaussian diffusion model with an identical architecture as our model which generates images \emph{without} the corresponding layouts. 
The visualization in Figure 8. of the main paper demonstrate that the internal features of this Gaussian-categorical diffusion model form distinct clusters compared to the Gaussian diffusion model. 

Second, we present a text-to-image generation approach that leverages semantic segmentation labels during training. 
Given text inputs, we sequentially generate layouts from texts and then images from the generated layouts. 
Specifically, we train a categorical diffusion model~\cite{d3pm} for text-to-layout generation and a layout-to-image synthesis model called SDM~\cite{sisdiff}. 
We train a modified version of SDM to incorporate text conditions to generate image from layouts. 

\begin{figure}[ht]
    \centering
    \hspace*{1.5cm}\includegraphics[width=0.7\textwidth]{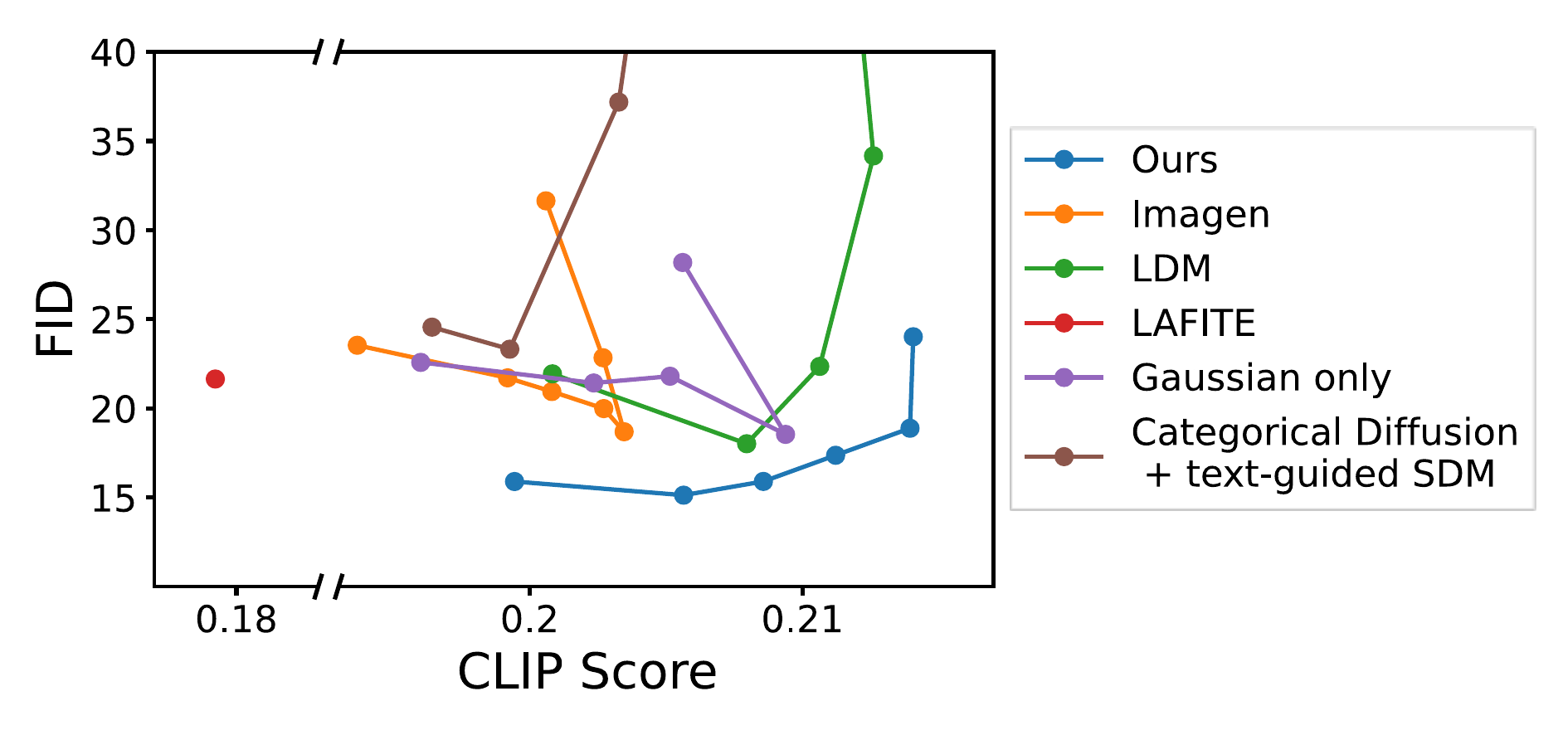}\hspace*{-1.5cm}
    \vspace{-0.1cm}
    \caption{FID-CLIP scores for the Gaussian diffusion on the MM CelebA-HQ-25 dataset, compared against existing approaches and the Gaussian-categorical diffusion.}
    \label{fig:ablation}
\end{figure}

To provide quantitative results, we report the FID-CLIP score pairs for the MM CelebA-HQ-25 in \Cref{fig:ablation}.
Our approach effectively enhances the performance of the Gaussian diffusion model by simultaneously generating corresponding semantic layouts. 
Also, our simultaneous generation of images and layouts outperforms the sequential generation from text to layouts and then to images. 

\section*{\LARGE A.8. Qualitative comparison}
\label{sec:quali}
\vspace{0.2cm}

\begin{figure}[b]
    \centering
    \includegraphics[width=\textwidth]{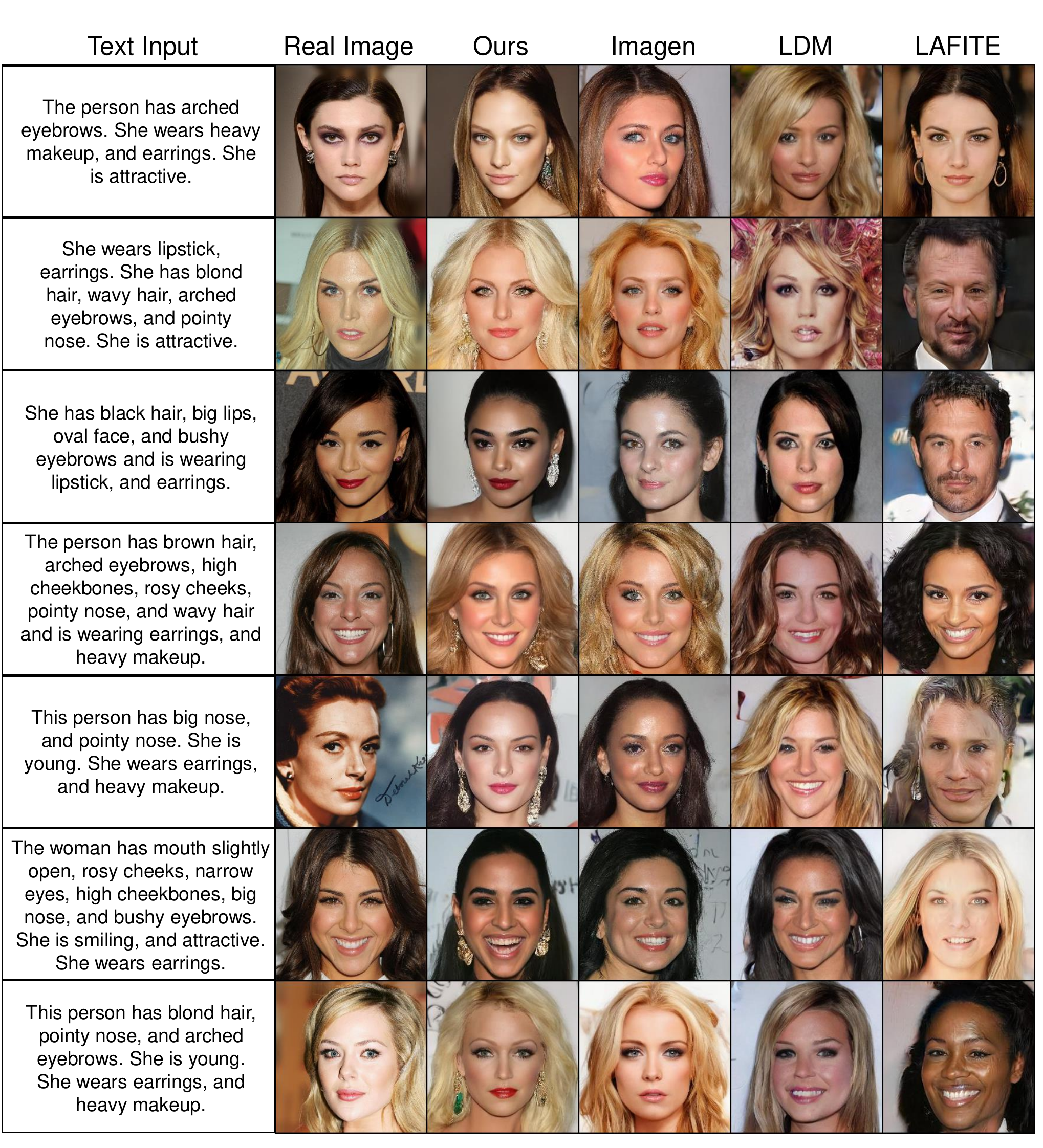}
    \caption{Qualitative comparison between the Gaussian-categorical diffusion model and existing text-to-image generation models on MM CelebA-HQ-25. We observe that existing models struggle to generate accessories such as earrings.}
    \label{fig:quali_celeba_earring}
\end{figure}

\begin{figure}[t]
    \centering
    \includegraphics[width=\textwidth]{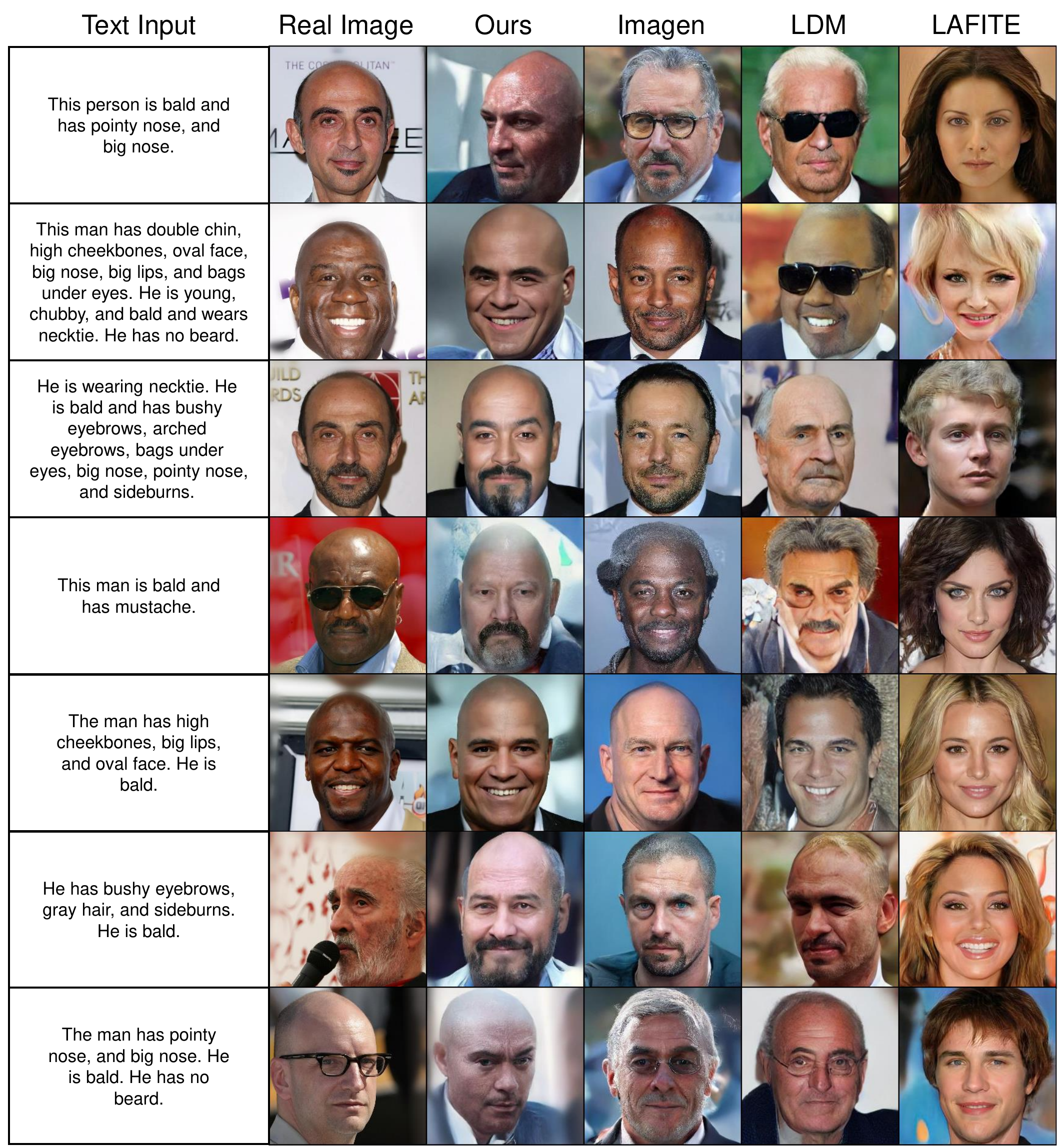}
    \caption{Qualitative comparison between the Gaussian-categorical diffusion model and existing text-to-image generation models on MM CelebA-HQ-25. We observe that existing models tend to generate hair even when given text conditions specifying baldness.}
    \label{fig:quali_celeba_bald}
\end{figure}

\begin{figure}[t]
    \centering
    \includegraphics[width=\textwidth]{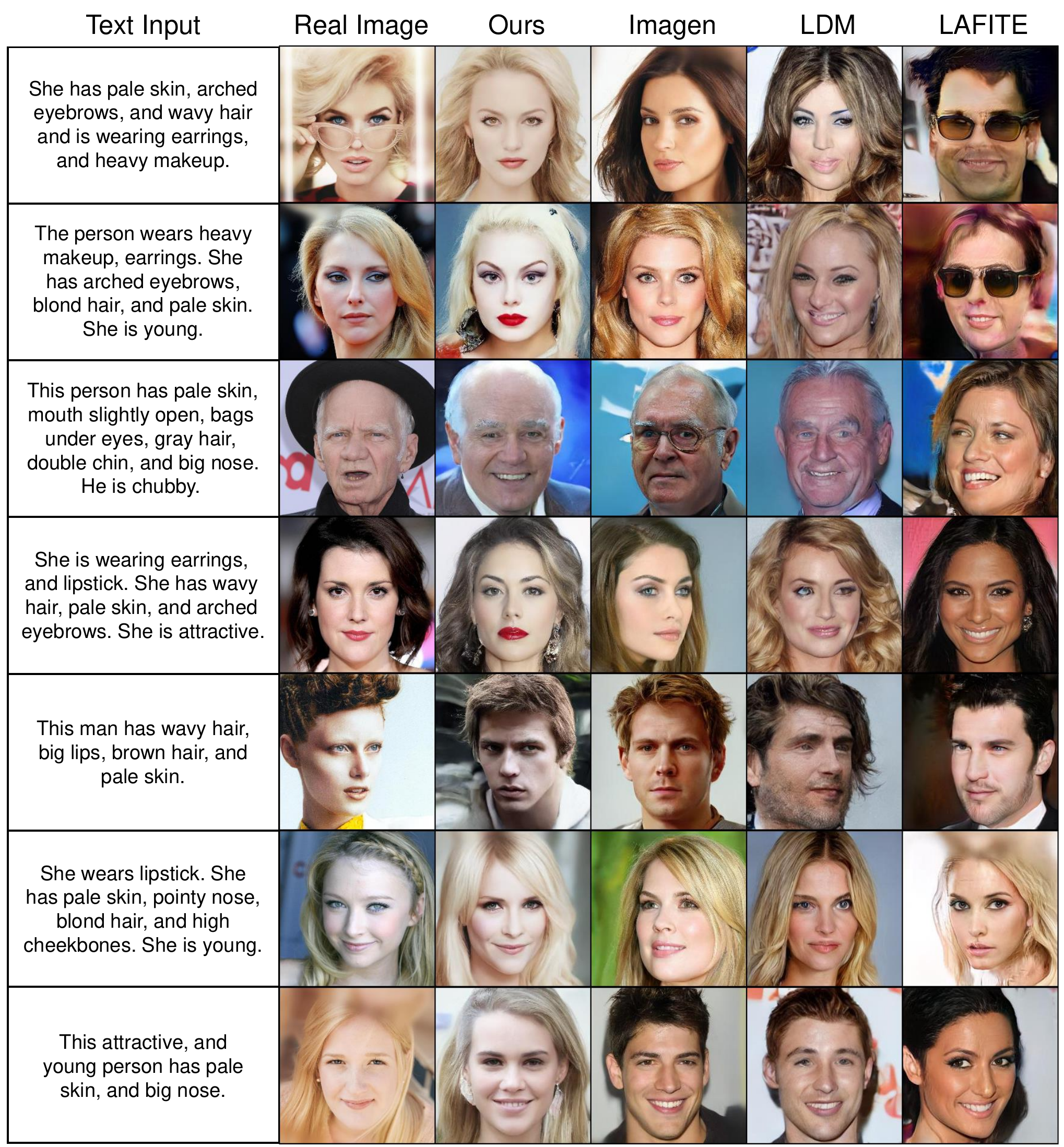}
    \caption{Qualitative comparison between the Gaussian-categorical diffusion model and existing text-to-image generation models on MM CelebA-HQ-25. We observe that existing approaches often fail to appropriately generate colors of skin.}
    \label{fig:quali_celeba_pale}
\end{figure}

\begin{figure}[t]
    \centering
    \includegraphics[width=\textwidth]{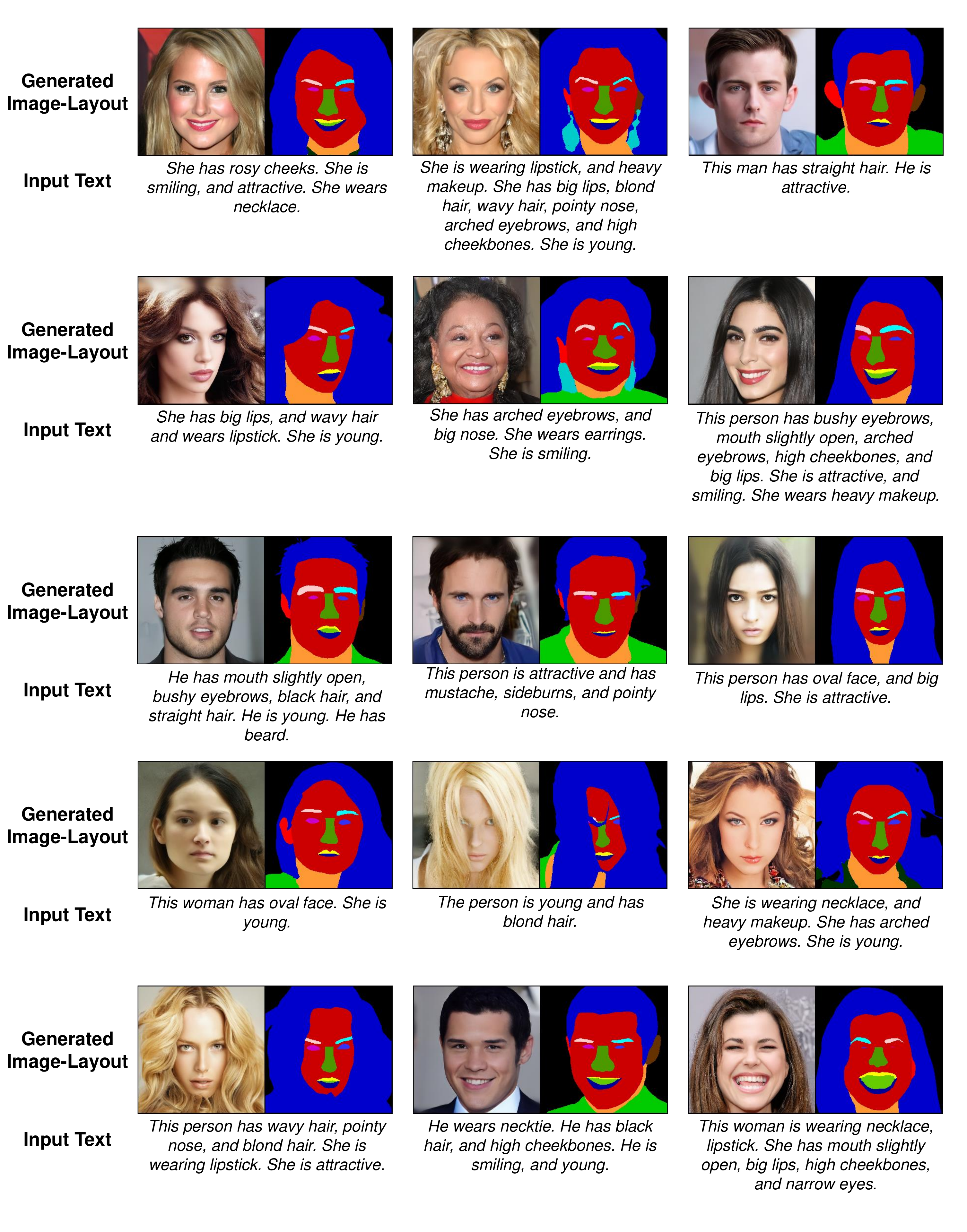}
    \vspace{-0.8cm}
    \caption{Example image-layout pairs generated by the Gaussian-categorical diffusion trained on the MM CelebA-HQ-100 dataset.}
    \label{fig:quali_ours_celeba}
\end{figure}

\begin{figure}[t]
    \centering
    \includegraphics[width=\textwidth]{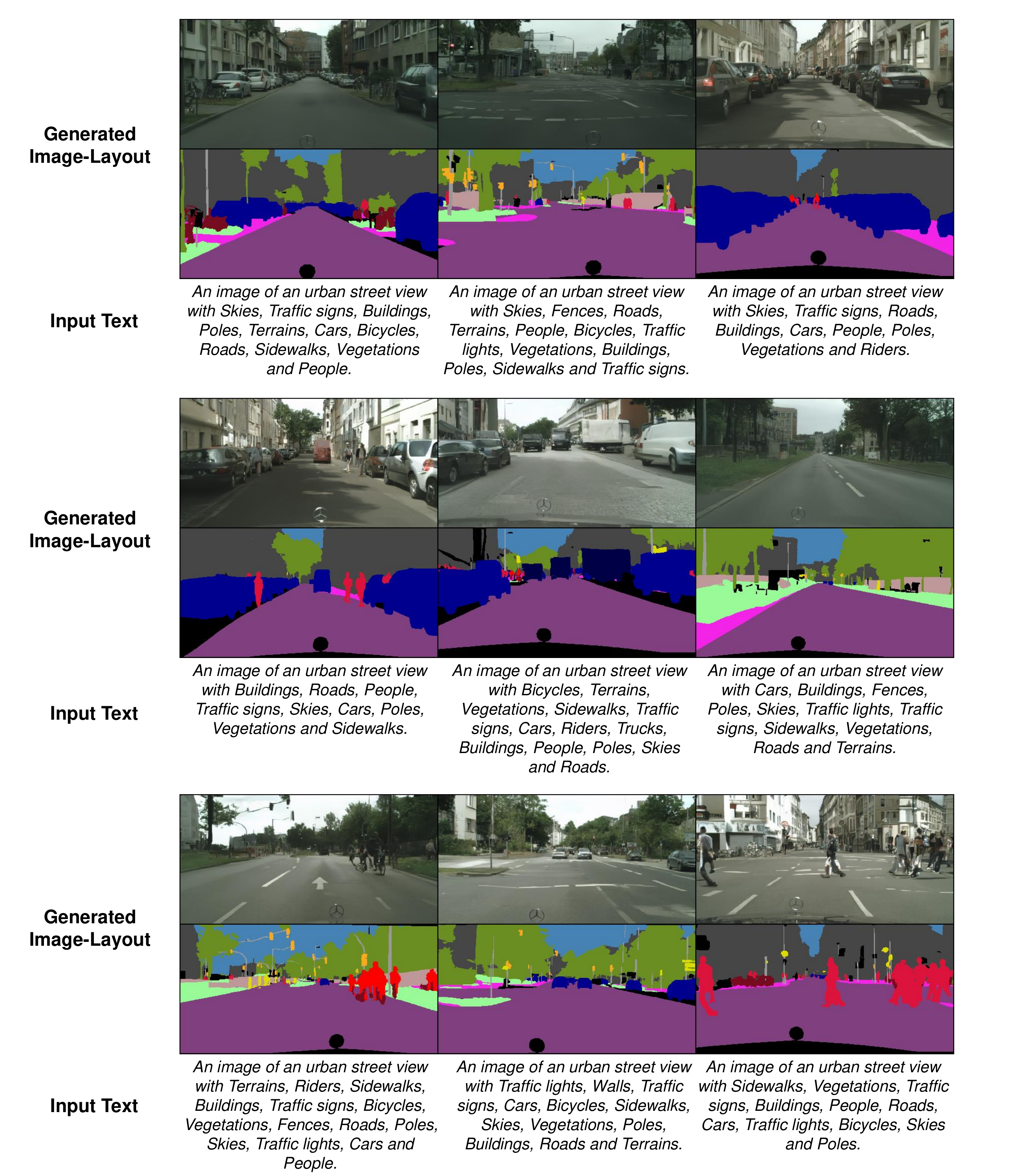}
    \caption{Example image-layout pairs generated by the Gaussian-categorical diffusion trained on the cityscapes dataset.}
    \label{fig:quali_ours_cityscapes}
\end{figure}

We provide the qualitative results from existing text-to-image generation models, and the Gaussian-categorical diffusion trained on MM CelebA-HQ-25 in the remaining supplementary material (\Cref{fig:quali_celeba_earring}, \Cref{fig:quali_celeba_bald}, and \Cref{fig:quali_celeba_pale}). 
Since diffusion-based models produce different results based on the guidance scale of the classifier-free guidance, we sample images from results exhibiting FID around 20. 
The guidance scales for each model to achieve an FID of 20 are 2, 10, and 10 for LDM, Imagen, and the Gaussian-categorical diffusion, respectively. 
We also provide uncurated results for generated image-layout pairs from the Gaussian-categorical diffusion model in \Cref{fig:quali_ours_celeba} and \Cref{fig:quali_ours_cityscapes}.
\vspace{-0.6cm}

\clearpage



\end{document}

%% file: math_commands.tex
\usepackage{amsmath,amsfonts,bm}

\newcommand{\brk}[1]{\left( #1 \right)}
\newcommand{\brkmid}[1]{\left\{ #1 \right\}}
\newcommand{\brkbig}[1]{\left[ #1 \right]}
\newcommand{\abs}[1]{\left\lvert#1\right\rvert}
\newcommand{\norm}[1]{\left\lVert#1\right\rVert}

\newcommand{\cond}[2]{\left(#1\,\middle|\,#2\right)}

\def\eqref#1{equation~\ref{#1}}

\def\1{\bm{1}}

\def\vmu{{\bm{\mu}}}

\def\mI{{\bm{I}}}

\def\mSigma{{\bm{\Sigma}}}
\def\mTheta{{\bm{\Theta}}}

\DeclareMathAlphabet{\mathsfit}{\encodingdefault}{\sfdefault}{m}{sl}
\SetMathAlphabet{\mathsfit}{bold}{\encodingdefault}{\sfdefault}{bx}{n}

\def\gC{{\mathcal{C}}}

\def\gN{{\mathcal{N}}}

\def\emY{{Y}}

\def\rvv{{\mathbf{v}}}

\def\rvx{{\mathbf{x}}}
\def\rvy{{\mathbf{y}}}
\def\rvz{{\mathbf{z}}}

\newcommand{\E}{\mathbb{E}}

\newcommand{\R}{\mathbb{R}}

